\documentclass[10pt,twocolumn,letterpaper]{article}

\usepackage[pagenumbers]{cvpr} 

\usepackage{graphicx}
\usepackage{amsmath}
\usepackage{amssymb}
\usepackage{booktabs}

\usepackage{color}
\usepackage{soul}
\usepackage{xcolor}
\usepackage{multirow}
\usepackage{booktabs}
\usepackage{enumerate}
\usepackage{enumitem}
\usepackage{pifont}%

\usepackage[pagebackref,breaklinks,colorlinks]{hyperref}

\usepackage[capitalize]{cleveref}
\crefname{section}{Sec.}{Secs.}
\Crefname{section}{Section}{Sections}
\Crefname{table}{Table}{Tables}
\crefname{table}{Tab.}{Tabs.}

\def\cvprPaperID{2339} 
\def\confName{CVPR}
\def\confYear{2023}

\begin{document}

\title{POTTER: Pooling Attention Transformer for Efficient Human Mesh Recovery}

\author{Ce Zheng$^{1}$\thanks{Work conducted during an internship at OPPO Seattle Research Center, USA.} , Xianpeng Liu$^{2}$, Guo-Jun Qi$^{3,4}$,  Chen Chen$^{1}$\\
$^1$Center for Research in Computer Vision, University of Central Florida\\
$^2$  North Carolina State University\\
$^3$  OPPO Seattle Research Center, USA \quad $^4$ Westlake University\\
{\tt\small cezheng@knights.ucf.edu; xliu59@ncsu.edu;  guojunq@gmail.com; chen.chen@crcv.ucf.edu}\\
}
\maketitle

\begin{abstract}
Transformer architectures have achieved SOTA performance on the human mesh recovery (HMR) from monocular images. However, the performance gain has come at the cost of substantial memory and computational overhead. A lightweight and efficient model to reconstruct accurate human mesh is needed for real-world applications. In this paper, we propose a pure transformer architecture named POoling aTtention TransformER (POTTER) for the HMR task from single images. Observing that the conventional attention module is memory and computationally expensive, we propose an efficient pooling attention module, which significantly reduces the memory and computational cost without sacrificing performance. Furthermore, we design a new transformer architecture by integrating a High-Resolution (HR) stream for the HMR task. The high-resolution local and global features from the HR stream can be utilized for recovering more accurate human mesh. Our POTTER outperforms the SOTA method METRO by only requiring 7\% of total parameters and 14\% of the Multiply-Accumulate Operations on the Human3.6M (PA-MPJPE metric) and 3DPW (all three metrics) datasets.  \textcolor{magenta}{The project
webpage is \url{https://zczcwh.github.io/potter_page/}}.           
\end{abstract}

\section{Introduction}
\label{sec:intro}

With the blooming of deep learning techniques in the computer vision community, rapid progress has been made in understanding humans from monocular images such as human pose estimation (HPE). No longer satisfied with detecting 2D or 3D human joints from monocular images, human mesh recovery (HMR) which can estimate 3D human pose and shape of the entire human body has drawn increasing attention. Various real-world applications such as gaming, human-computer interaction, and virtual reality (VR) can be facilitated by HMR with rich human body information. However, HMR from single images is extremely challenging due to  complex human body articulation, occlusion, and depth ambiguity.

\begin{figure}[]
\vspace{-10pt}
  \centering
  \includegraphics[width=0.70\linewidth]{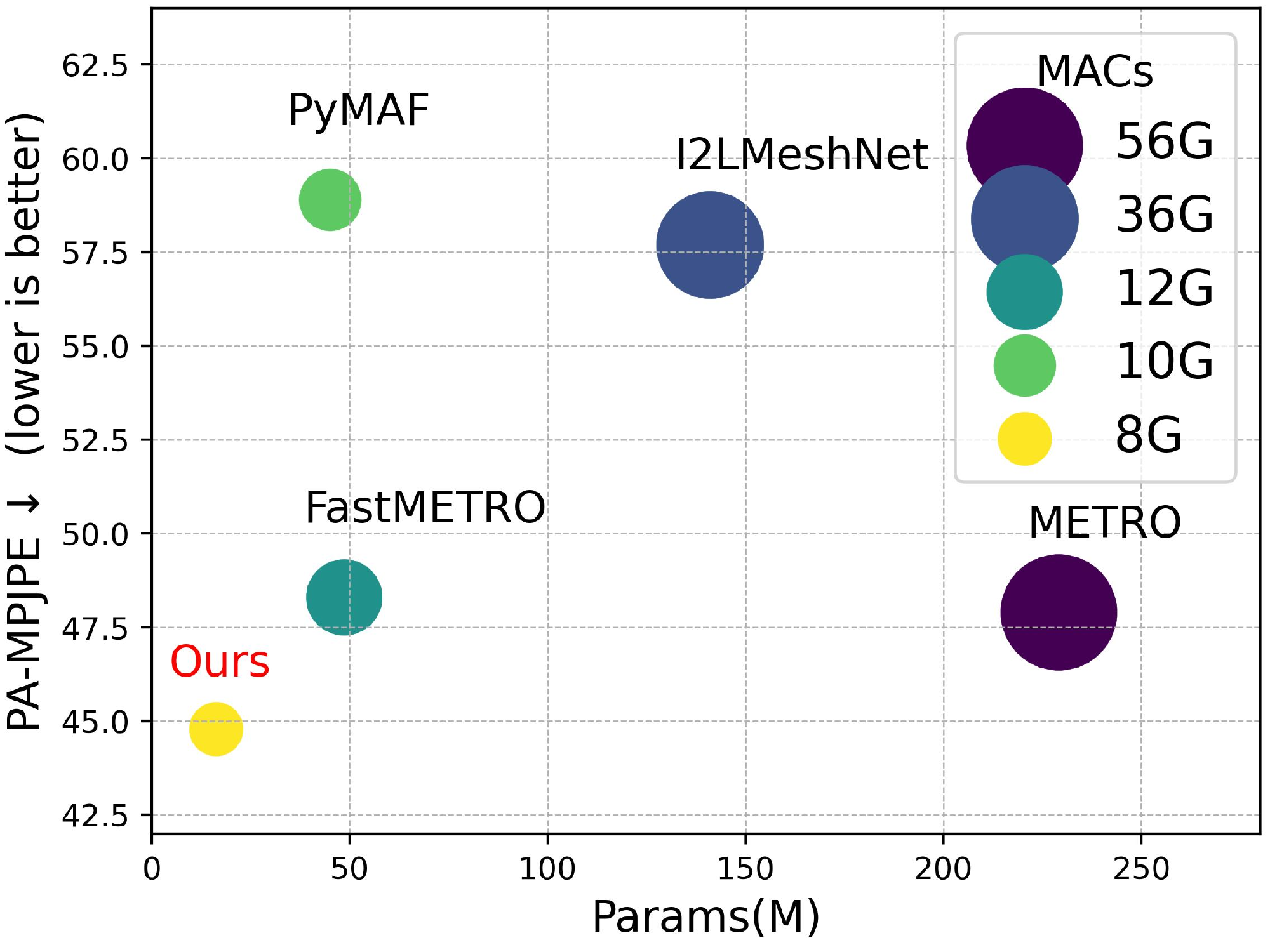}
  \vspace{-10pt}
  \caption{HMR performance comparison with Params and MACs on 3DPW dataset. We outperform SOTA methods METRO\cite{lin2021metro} and FastMETRO\cite{FastMETRO} with much fewer Params and MACs. PA-MPJPE is the Procrustes Alignment Mean Per Joint Position Error.}
  \label{fig:macs_com}
  \vspace{-15pt}
\end{figure}

Recently, motivated by the evolution of the transformer architecture in natural language processing, Vision Transformer (ViT) \cite{Dosovitskiy2020ViT} successfully introduced transformer architecture to the field of computer vision. The attention mechanism in transformer architecture demonstrates a strong ability to model global dependencies in comparison to the Convolutional Neural Network (CNN) architecture. With this trend, the transformer-based models have sparked a variety of computer vision tasks, including object detection\cite{misra2021end,liu2021group}, semantic segmentation \cite{xie2021segformer,maskformer}, and video understanding \cite{liu2021video,videomae} with promising results. For HMR, the SOTA methods \cite{lin2021metro,FastMETRO} all utilize the transformer architecture to exploit non-local relations among different human body parts for achieving impressive performance.

However, one significant limitation of these SOTA HMR methods is model efficiency. Although transformer-based methods \cite{lin2021metro,lin2021_mesh_graphormer} lead to great improvement in terms of accuracy, the performance gain comes at the cost of a substantial computational and memory overhead. The large CNN backbones are needed for \cite{lin2021metro,lin2021_mesh_graphormer} to extract features first. Then, computational and memory expensive transformer architectures are applied to process the extracted features for the mesh reconstruction. Mainly pursuing higher accuracy is not an optimal solution for deploying HMR models in real-world applications such as human-computer interaction, animated avatars, and VR gaming (for instance, SOTA method METRO \cite{lin2021metro} requires 229M Params and 56.6G MACs as shown in Fig. \ref{fig:macs_com}). Therefore, it is important to also consider the memory footprint and computational complexity when evaluating an HMR model.


\begin{figure}[htp]
\vspace{-5pt}
  \centering
  \includegraphics[width=1\linewidth]{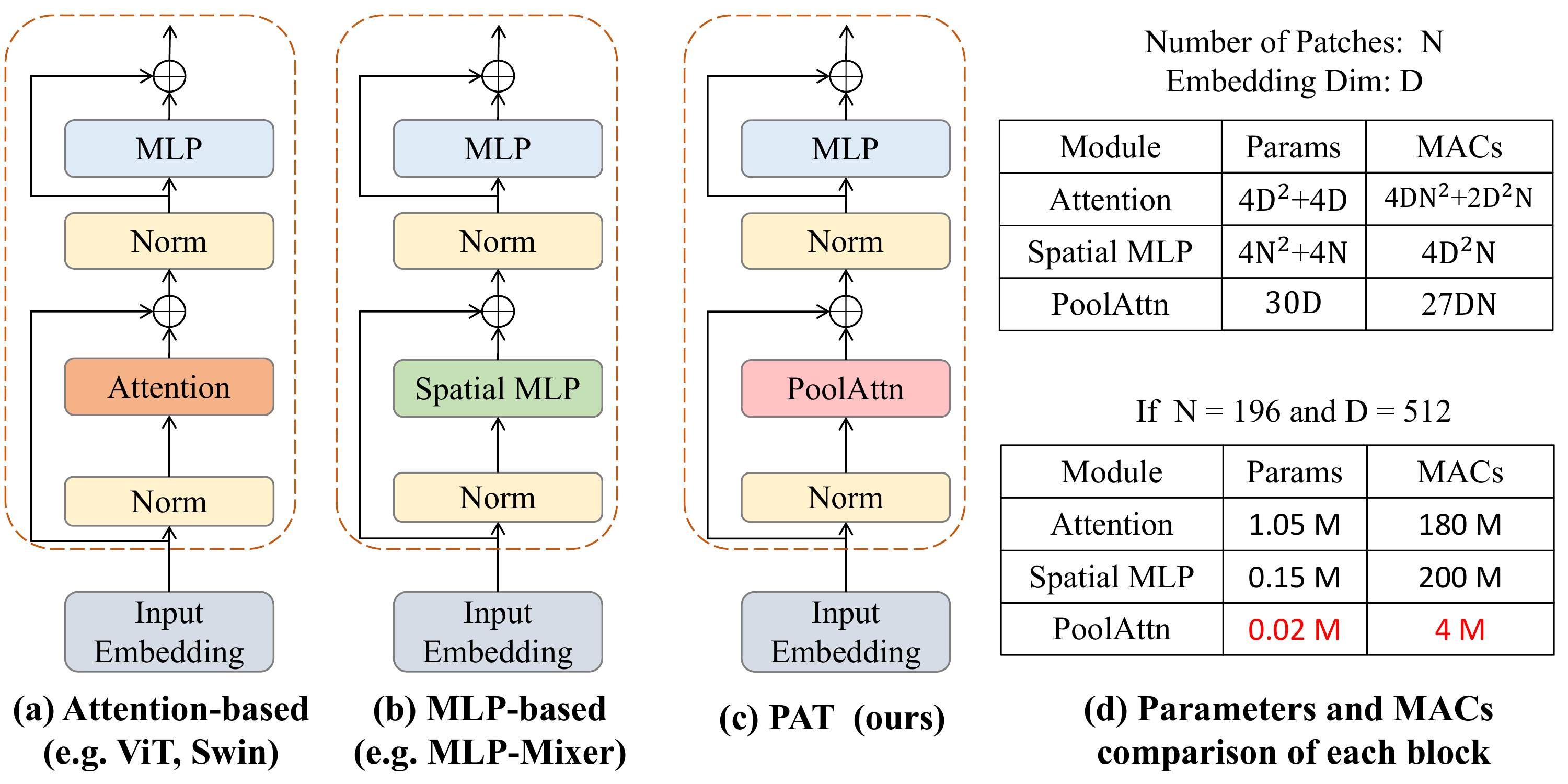}
  \vspace{-15pt}
  \caption{Transformer blocks of different models. We suppose the number of patches (N) and the embedding dimension (D) for each block are the same when comparing the Params and MACs. 
  }
  \label{fig:block_com}
  \vspace{-10pt}
\end{figure}

To bridge this gap, we aim to design a lightweight end-to-end transformer-based network for efficient HMR. Observing that the transformer blocks (attention-based approaches in Fig. \ref{fig:block_com} (a) and MLP-based approaches in Fig. \ref{fig:block_com} (b)) are usually computational and memory consuming, we 
propose a Pooling Attention Transformer (PAT) block as shown in Fig. \ref{fig:block_com} (c) to achieve model efficiency. 
After patch embedding, the image input becomes $X = [D,\frac{H}{p},\frac{W}{p}]$, where $D$ is the embedding dimension and the number of patches is $N = \frac{H}{p} \times \frac{W}{p}$ when patch size is $ p \times p$. The input for transformer block is often written as $X_{in} = [N, D]$. 
To reduce the memory and computational costs, we design a Pooling Attention (PoolAttn) module in our PAT block. The PoolAttn consists of patch-wise pooling attention and embed-wise pooling attention. For the patch-wise pooling attention block, we preserve the patches' spatial structure based on the input $X_{in} = [D,\frac{H}{p},\frac{W}{p}]$, then apply patch-wise pooling attention to capture the correlation of all the patches. For the embed-wise pooling attention block, we maintain the 2D spatial structure of each patch (without flattening to 1D  embedded features). The input is reshaped to $X_{in} = [N,D_h,D_w]$, where $D_h \times D_w = D$ is the embedding dimension. The embed-wise pooling attention is applied to model the dependencies of the embedding dimensions in each patch. A detailed explanation is provided in Section \ref{sec:PoolAttnDesign}. The Params and MACs comparison between the PoolAttn and conventional attention module or MLP-based module is shown in Fig. \ref{fig:block_com} (d). Thus, PAT can reduce the Params and MACs significantly while maintaining high performance, which can be utilized for efficient HMR.

\begin{figure}[]
\vspace{-5pt}
  \centering
  \includegraphics[width=1.\linewidth]{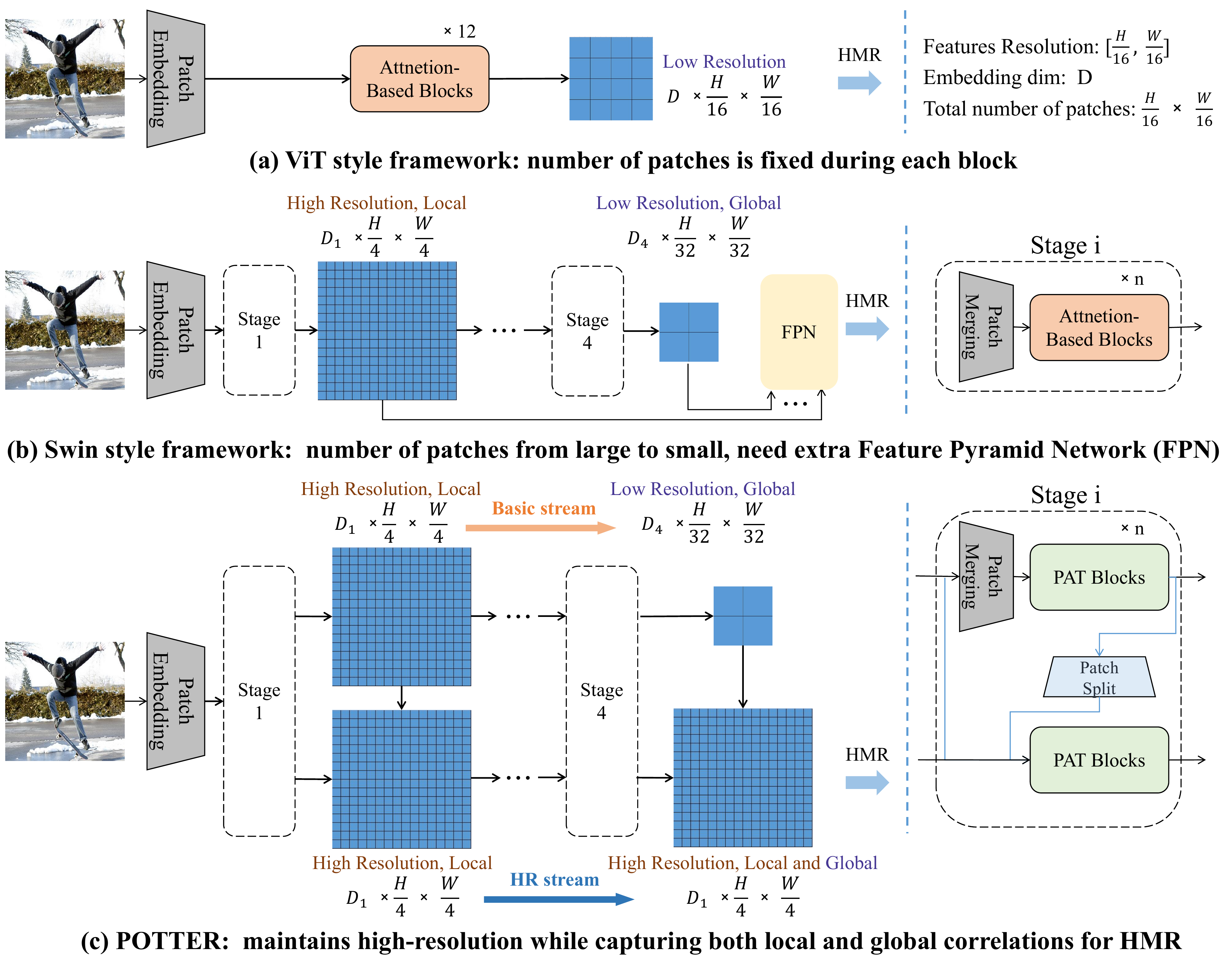}
  \vspace{-20pt}
  \caption{The illustration in terms of patches during each stage in transformer architectures.}
  \label{fig:patch_com}
  \vspace{-15pt}
\end{figure}

Equipped with PAT as our transformer block, the next step for building an efficient and powerful transformer-based HMR model is to design an overall architecture. The naive approach is to apply a Vision Transformer \cite{Dosovitskiy2020ViT} (ViT) architecture as shown in Fig. \ref{fig:patch_com} (a). The image is first split into patches. After patch embedding, a sequence of patches is treated as tokens for transformer blocks. But in ViT, patches are always within a fixed scale in transformer blocks, producing low-resolution features. For the HMR task, high-resolution features are needed because human body parts can vary substantially in scale. Moreover, ViT architecture focuses on capturing the global correlation, but the local relations can not be well modeled. Recently, Swin \cite{liu2021Swin} introduced a hierarchical transformer-based architecture as shown in Fig. \ref{fig:patch_com} (b). It has the flexibility to model the patches at various scales, the global correlation can be enhanced during hierarchical blocks. However, it also produces low-resolution features after the final stage. To obtain high-resolution features, additional CNN networks such as Feature Pyramid Network \cite{lin2017feature} (FPN) are required to aggregate hierarchical feature maps for HMR. Thus, we propose our end-to-end architecture as shown in Fig. \ref{fig:patch_com} (c), the hierarchical patch representation ensures the self-attention can be modeled globally through transformer blocks with patch merge. To overcome the issue that high-resolution representation becomes low-resolution after patch merge, we propose a High-Resolution (HR) stream that can maintain high-resolution representation through patch split by leveraging the local and global features from the basic stream. Finally, the high-resolution local and global features are used for reconstructing accurate human mesh. The entire framework is also lightweight and efficient by applying our PAT block as the transformer block.

Our contributions are summarized as follows: 
\begin{itemize}[noitemsep,leftmargin=*]  

\item We propose a Pooling Transformer Block (PAT) which is composed of the Pooling Attention (PoolAttn) module to reduce the memory and computational burden without sacrificing performance.

\item We design a new transformer architecture for HMR by integrating a High-Resolution (HR) stream. Considering the patch's merging and split properties in transformer, the HR stream returns high-resolution local and global features for reconstructing accurate human mesh.  

\item Extensive experiments demonstrate the effectiveness and efficiency of our method -- POTTER. In the HMR task, POTTER surpasses the transformer-based SOTA method METRO\cite{lin2021metro} on Human3.6M (PA-MPJPE metric) and 3DPW (all three metrics) datasets with only 7 \% of Params and 14 \% MACs.
\end{itemize}

\begin{figure*}[htp]
  \centering
  \includegraphics[width=0.91\linewidth]{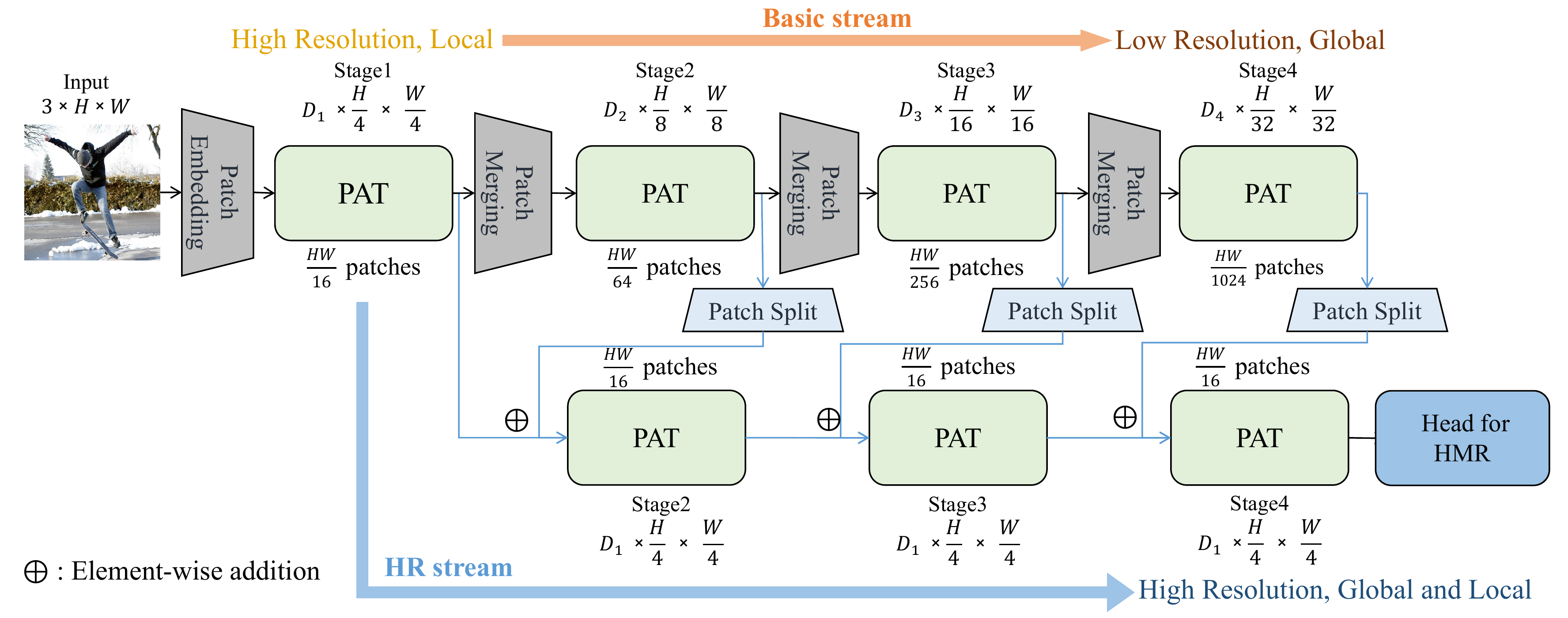}
  \vspace{-10pt}
  \caption{The overall architecture of our POTTER. PAT is our proposed Pooling Attention Transformer block. The basic stream of POTTER adopts hierarchical architecture with 4 stages \cite{liu2021Swin}, where the number of patches is gradually reduced for capturing more global information with low-resolution features ($\frac{H}{32} \times \frac{W}{32}$). Our proposed HR stream maintains the high-resolution ($\frac{H}{4} \times \frac{W}{4}$) feature representation at each stage. The global features from the basic stream are fused with the local features by patch split blocks in the HR stream. Thus, the high-resolution local and global features are utilized for the HMR task. }
  \label{fig:main_arch}
  \vspace{-15pt}
\end{figure*}

\section{Related Work}
Since the HMR is one of the fundamental tasks in computer vision with a long history, here we focus on the more recent and relevant approaches. Readers can explore more detailed information about HMR in a recent and comprehensive HMR survey \cite{hmrsurvey}. 

\textbf{HMR from a single image.}
HMR has attracted increasing attention in recent years. 
Most of the HMR methods ~\cite{pymaf2021,kanazawaHMR18,Jiang_2020_CVPR,xu2020Low_Resolution,Thundr} utilize a parametric human model such as SMPL \cite{SMPL:2015} to reconstruct human mesh by estimating the pose and shape parameters of the parametric model. 
Kolotouros et al. \cite{kolotouros2019cmr} present a graph convolution neural network to learn the vertex-vertex relations. They regress the 3D mesh vertices directly instead of the SMPL model parameters. SPIN ~\cite{Kolotouros2019SPIN} combines the regression and optimization process in a loop. The regressed output served as better initialization for optimization (SMPLify). Similarly, PyMAF \cite{pymaf2021} propose a pyramidal mesh alignment feedback where the mesh-aligned evidence is exploited to correct parametric errors. ProHMR \cite{prohmr} is a probabilistic model that outputs a conditional probability distribution using conditional normalizing flow. Dwivedi et al. \cite{dsr2021} propose a differentiable semantic rendering loss to exploit richer image information about clothed people.

\textbf{Transformers in HMR.}
Transformers are first proposed by \cite{vaswani2017attention} in the field of NLP. Inspired by the success of transformer's attention mechanism when dealing with token input, many researchers apply the transformer in various vision tasks such as object detection ~\cite{carion2020end,zhu2020deformable}, image classification ~\cite{Dosovitskiy2020ViT,liu2021Swin}, segmentation ~\cite{zheng2021rethinking}, human pose estimation ~\cite{Poseformer_2021_ICCV,zhao2021graformer}, etc. METRO \cite{lin2021metro} is the first transformer-based method for HMR. After extracting the image features by CNN backbone, a transformer encoder is proposed to model vertex-vertex and vertex-joint interaction. Although METRO outperforms the previous SOTA methods by a large margin (more than 10 MPJPE on Human3.6M and 3DPW datasets), METRO requires substantial memory and computational costs to achieve this impressive performance. As an extended version of METRO, MeshGraphormer \cite{lin2021_mesh_graphormer} further combines the graph convolutional network (GCN) with a transformer to model local and global interactions among mesh vertices and joints. It still incurs substantial memory and computational overhead. Zeng et al. \cite{tcformer} propose a Token Clustering Transformer (TCFormer) to merge tokens from different locations by progressive clustering procedure. However, the performance of TCFormer can not beat METRO and MeshGraphormer.

\textbf{Efficient models for HMR.}
As mentioned above, \cite{lin2021metro,lin2021_mesh_graphormer} pursue higher accuracy by sacrificing computational and memory efficiency. For real-world applications, model efficiency is also a key metric when evaluating HMR models, while less studied before. Although FastMETRO \cite{FastMETRO} reduces the computational and memory costs for the transformer part, it still relies on the heavy CNN backbone to achieve impressive performance. Another attempt is to reconstruct human mesh from a 2D human pose, which is proposed by GTRS \cite{gtrs}. A lightweight transformer model employing a lightweight 2D pose detector can reduce computational and memory costs significantly. 
However, the performance of GTRS is not comparable to the SOTA methods since it only uses the 2D pose as input, some information such as human shape is missing.

\section{Methodology}
\subsection{Overall Architecture}
We propose an end-to-end transformer network named POTTER for the HMR task as shown in Fig. \ref{fig:main_arch}. Following the general hierarchical transformer architecture (such as Swin \cite{liu2021Swin}), there are four hierarchical stages of transformer blocks. After the patch embedding, the input image  $X_{img} \in \mathbb{R} ^{3 \times H \times W}$ is embedded to the input features of ``stage 1'' $X_{in1} \in \mathbb{R} ^{D_{1} \times \frac{H}{4} \times \frac{W}{4}}$, where $D_{1}$ is the embedding dimension of ``stage 1'', $H$ and $W$ are the height and width of the input image, respectively. The total number of patches is $\frac{H}{4} \times \frac{W}{4} = \frac{HW}{16}$ and the resolution of the features is $[\frac{H}{4}, \frac{W}{4}]$. After the transformer blocks modeling the attention, the output $X_{out1}$ keeps the same size as the input $X_{in1} \in \mathbb{R} ^{D_{1} \times \frac{H}{4} \times \frac{W}{4}}$. 

In the basic stream, we follow the Swin \cite{liu2021Swin} style  hierarchical architecture, a patch merging block is applied between two adjacent stages to build hierarchical feature maps, which reduces the number of patches between two adjacent stages. Thus, the output of ``stage 2'' becomes  $X^{basic}_{out2} \in \mathbb{R} ^{D_{2} \times \frac{H}{8} \times \frac{W}{8}}$, the total number of patches is reduced to  $\frac{H}{8} \times \frac{W}{8} = \frac{HW}{64}$, and the resolution of the features is decreased to $[\frac{H}{8}, \frac{W}{8}]$. This procedure is the same for ``stage 3'' and ``stage 4'', where the output is $X^{basic}_{out3} \in \mathbb{R} ^{D_{3} \times \frac{H}{16} \times \frac{W}{16}}$ and $X^{basic}_{out4} \in \mathbb{R} ^{D_{4} \times \frac{H}{32} \times \frac{W}{32}}$, respectively. 

In the High-Resolution (HR) stream, a patch split block is applied between the basic stream and the HR steam, which splits the merged patches to maintain a high-resolution feature representation. Thus, the output for ``stage 3'' and ``stage 4'' are $X^{HR}_{out3} \in \mathbb{R} ^{D_{1} \times \frac{H}{4} \times \frac{W}{4}}$ and $X^{HR}_{out4} \in \mathbb{R} ^{D_{1} \times \frac{H}{4} \times \frac{W}{4}}$, respectively. The total number of patches during the HR stream is kept as $\frac{H}{4} \times \frac{W}{4} = \frac{HW}{16}$ and the resolution of the features is maintained as $[\frac{H}{4}, \frac{W}{4}]$.

\subsection{The Design of Pooling Attention}
\label{sec:PoolAttnDesign}

\begin{figure}[htp]
\vspace{-5pt}
  \centering
  \includegraphics[width=1\linewidth]{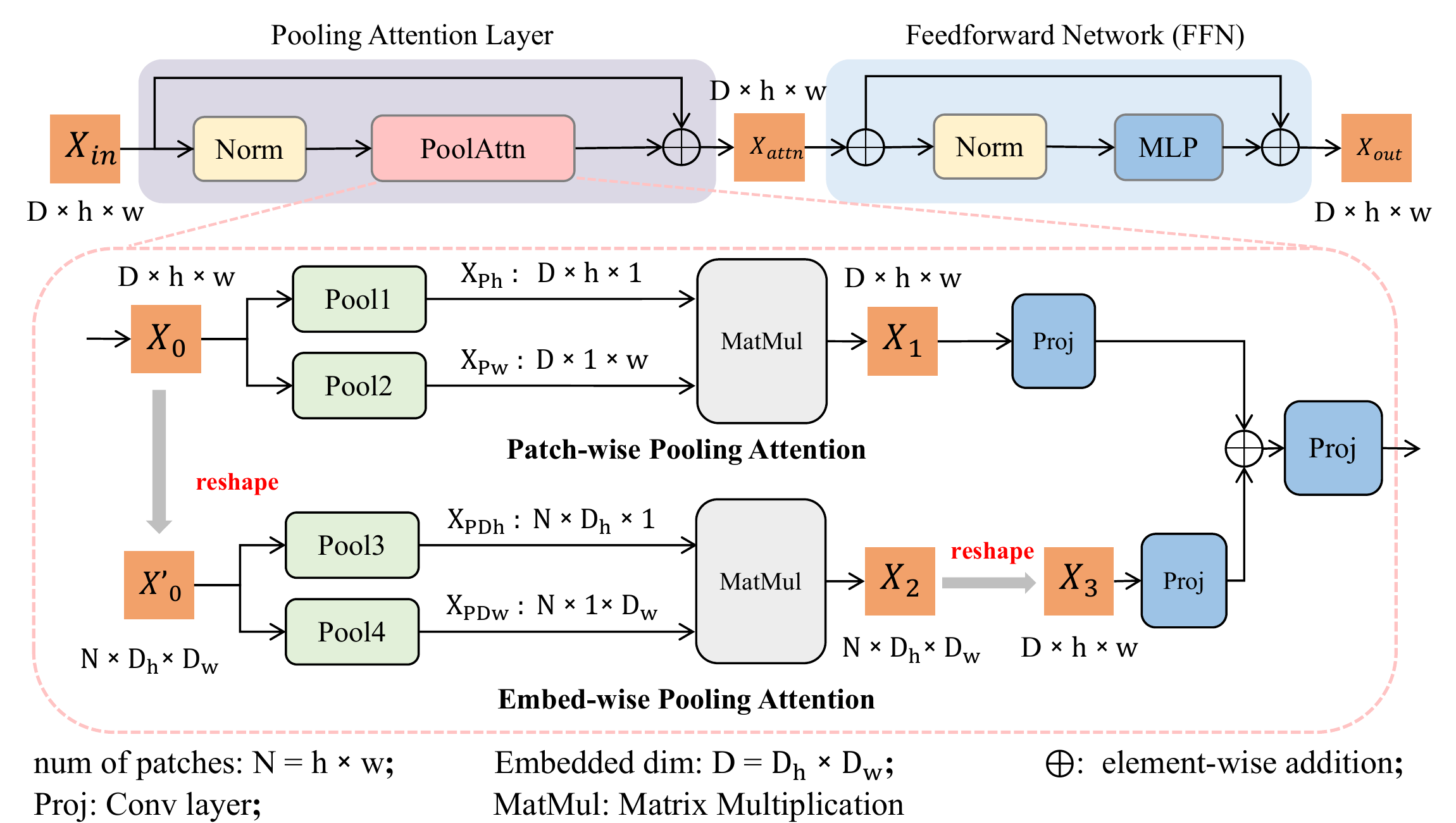}
  \vspace{-20pt}
  \caption{The Pooling Attention Transformer (PAT) block}
  \label{fig:pool_attn}
  \vspace{-5pt}
\end{figure}

Following the conventional transformer models, our Pooling Attention Transformer (PAT) block has a similar structure as shown in Fig. \ref{fig:block_com}. Among various transformer-based architecture, PoolFormer block \cite{poolformer} is one of the closely related works to our PoolAtten block. In PoolFormer \cite{poolformer}, a simple spatial pooling layer is used to replace the attention for patch mixing. Specifically, given the input $X_{in} \in \mathbb{R} ^{D \times h \times w}$ where $D$ is the embedding dimension and $h \times w$ is the number of patches, an average 2D pooling layer ($pool\_size = 3 \times 3, stride = 1, padding = 1$) is applied as patch mixer. The output  $X_{out} \in \mathbb{R} ^{D \times h \times w}$ keeps the same size as the input.  Surprisingly, the performance of PoolFormer surpasses many complicated transformer-based models with much less computational and memory complexity. 

Different from PoolFormer, we apply the pooling attention module to model the patch-wise attention and embed-wise attention while reducing computational and memory costs. 
The detailed structure of our PAT block is shown in Fig. \ref{fig:pool_attn}. The Pooling Attention (\textbf{PoolAttn}) consists of patch-wise pooling attention and embedding dimension-wise pooling attention (embed-wise for short). Given the input $X_{in} \in \mathbb{R} ^{D \times h \times w}$, where $D$ is the embedding dimension and $h \times w$ is the number of patches, we first apply a Layer Normalization (LN) operation to normalize the input $X_{in}$ as $X_{0}$, then, a PoolAttn module is used for calculating the attention based on the squeezed 2D features.

\begin{figure*}[htp]
  \centering
  \includegraphics[width=0.85\linewidth]{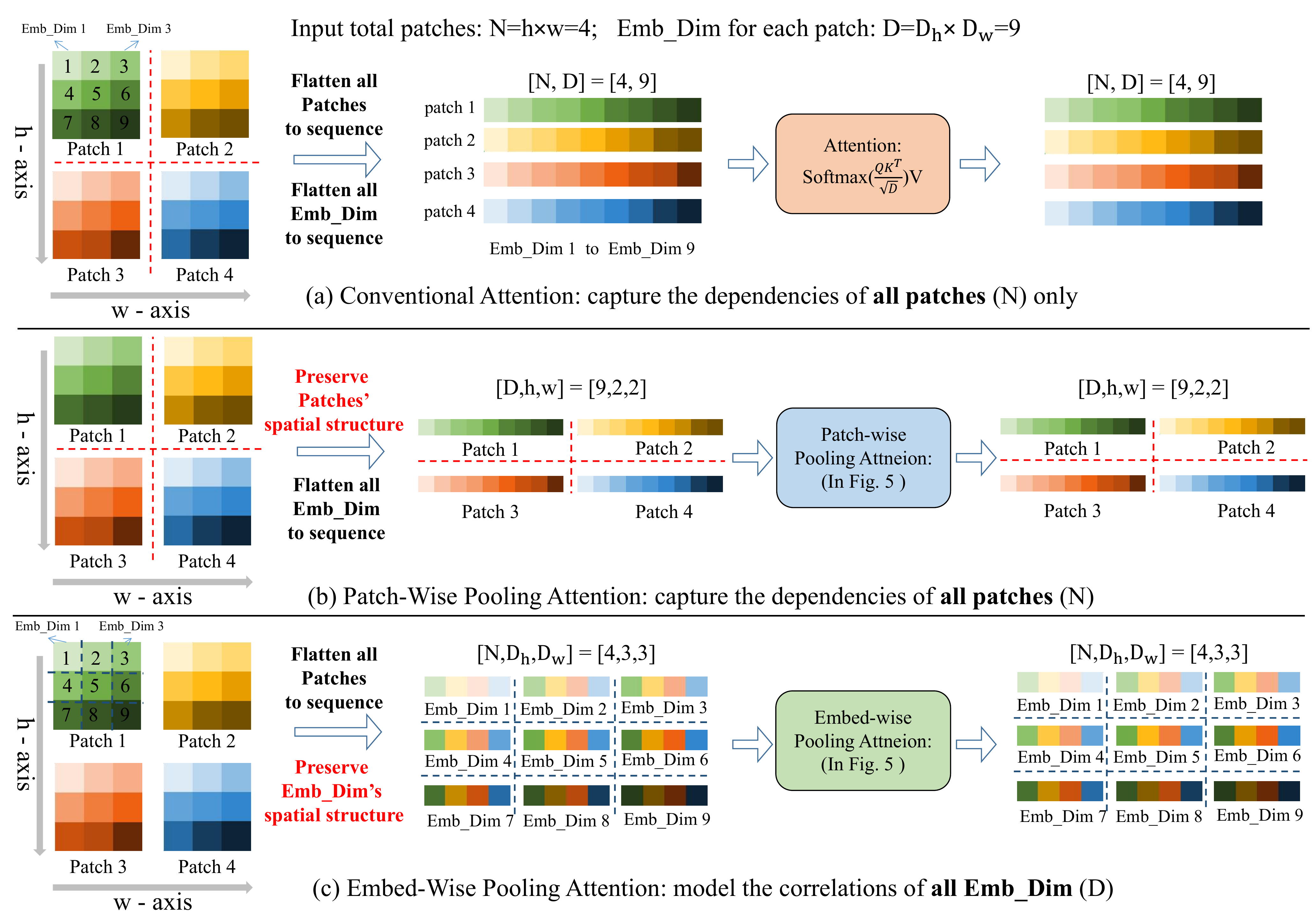}
  \vspace{-10pt}
  \caption{The illustration of the patch-wise pooling attention and the embed-wise pooling attention. The conventional attention blocks not only reorder all the patches as a sequence, but also flatten the embedded features of each patch to 1D features. For patch-wise attention, we preserve the patch's spatial structure, then apply pooling attention to capture the correlations between all the patches. For the embed-wise attention, we maintain the spatial structure of 2D embedded features for each patch, then apply pooling attention to model the correlations between the embedding dimensions.      }
  \label{fig:pool_attn_exp}
  \vspace{-10pt}
\end{figure*}

A graphical illustration of the procedures of the patch-wise pooling attention and the embed-wise pooling attention is presented in Fig. \ref{fig:pool_attn_exp}. We suppose the number of patches is $h \times w = 2 \times 2$ and the embedding dimension for each patch is $D=3 \times 3=9$ in this simple illustration. \textit{Unlike the conventional attention block that all the patches are reordered as a sequence and the embedded features of each patch are flattened, we preserve the patch's spatial structure before applying the patch-wise pooling attention. Similarly, for each patch, we maintain the 2D spatial embedded features representation (without flattening to 1D features) before applying embed-wise pooling attention.}   

For the \textbf{patch-wise pooling attention}, we squeeze the $X_{0}$ along the $h-axis$ and $w-axis$ by two average pooling layers ($Pool_1$ and $Pool_2$), returning the $X_{Ph} \in \mathbb{R} ^{D \times h \times 1}$  and the $X_{Pw} \in \mathbb{R} ^{D \times 1 \times w}$. The matrix multiplication (MatMul) results of these 2D features $X_{Ph}$ and  $X_{Pw}$ is the patch attention, named  $X_{1} \in \mathbb{R} ^{D \times h \times w} $.      
\begin{align}
\small
    & X_{Ph} = Pool_1(X_0), \quad  X_{Ph} \in \mathbb{R} ^{D \times h \times 1} \\
    & X_{Pw} = Pool_2(X_0), \quad  X_{Pw} \in \mathbb{R} ^{D \times 1 \times w} \\
    & X_1 = MatMul( X_{ph} , X_{pw})
\end{align}

Previous transformer-based methods only model the attention given the number of patches. We further exploit the correlation given the embedding dimensions. For the \textbf{embed-wise pooling attention}, we reshape the $X_{0} \in \mathbb{R} ^{D \times h \times w}$ as the  $X'_{0} \in \mathbb{R} ^{N \times D_h \times D_w}$, where $N=h \times w$ and $D = D_h \times D_w$. Similarly, we squeeze the $X'_{0}$ along the $D_h-axis$ and $D_w-axis$ by two average pooling layers ($Pool_3$ and $Pool_4$). The squeezed 2D features $X_{PDh} \in \mathbb{R} ^{N \times D_h \times 1}$  and the $X_{PDw} \in \mathbb{R} ^{N \times 1 \times D_w}$ are used for compute the embedding attention $X_{2} \in \mathbb{R} ^{N \times D_h \times D_w} $ by the matrix multiplication. 
\begin{align}
\small
    & X_{PDh} = Pool_3(X'_0), \quad  X_{PDh} \in \mathbb{R} ^{N \times D_h \times 1} \\
    & X_{PDw} = Pool_4(X'_0), \quad  X_{PDw} \in \mathbb{R} ^{N \times 1 \times D_w} \\
    & X_2 = MatMul( X_{PDh} , X_{PDw})
\end{align}

Next, the embedding attention $X_{2}$ is reshaped back to $X_{3} \in \mathbb{R} ^{D \times h \times w}$. A projection layer (implemented by a CONV layer, same as in \cite{poolformer}) projects the sum of patch attention $X_{1}$ and the embedding attention $X_{3}$ as the PoolAttn's output. 
\begin{align}
\small
    X_3^{'} = Proj_3(Proj_1(X_{1}) + Proj_2(X_{3}))
\end{align}

Thus, the patch-wise pooling attention preserves the patch's spatial locations when capturing the correlations between all the patches. In the meantime, the embed-wise pooling attention maintains the spatial embedded features representation for each patch when modeling the correspondences of embedded features. Compared with the simple pooling module in PoolFormer \cite{poolformer}, the PoolAttn module boosts the performance by a large margin without increasing memory and computational cost which is verified in Section \ref{sec:Ablation}.

With the PoolAttn module, one PAT block returns the output $X_{out}$ given the block's input $X_{in}$, and can be expressed as: 
\begin{align}
\small
    & X_{attn} = PoolAttn(LN(X_{in})) + X_{in} \\
    & X_{out} = MLP(LN(X_{attn})) + X_{attn}
\end{align}
Our PoolAttn operation significantly reduces the computational and memory costs of the PAT. The detailed parameters and MACs comparison of one PoolAttn module is shown in the \textcolor{blue}{\textcolor{blue}{Supplementary \ref{supp_Complexity}}}. For instance, the input is with the shape of $ [512,16,16] $ where the embedding dimension is 512 and the number of patches is $16 \times 16 = 196$. One attention block such as in ViT or Swin requires 1.1M params and 180M MACs, while our PoolAttn only requires 0.02M params (2\%) and 4M MACs (2\%) without downgrading the performance.  

\subsection{High-Resolution Stream}

Different from the general hierarchical transformer architecture, the output of ``stage 1'' $X_{out1}$ is also used by the HR streams as shown in Fig .\ref{fig:main_arch}. 
For the HR stream, we aim to maintain the high-resolution features which are critical for the HMR task. Thus, we do not need a patch merging layer to merge the patches, which downgrades the high-resolution representation to the low-resolution representation. Instead, we aggregate the information from the basic stream to the HR stream. The patch split layer can convert the merged patches $X_{out2}^{basic}$ back to the high-resolution patches. 


During the basic stream (which is the architecture of Swin \cite{liu2021Swin} equipped with our PAT block), the high-resolution and local features gradually become the low-resolution and global features. Thus, it is served as the general vision backbone (usually for the classification task). To reconstruct human mesh, global features from the basic stream are aggregated with the local features in the HR stream, returning high-resolution local and global features. The input of the transformer block of each stage during the HR stream can be expressed as:
\begin{align}
\small
    & X_{out \ i+1}^{HR} = PatchSplit(X_{out \ i}^{basic}) + X_{out \ i}^{HR}  
\end{align}

\noindent Thus, the output for the $i$ stage is $X^{HR}_{out \ i} \in \mathbb{R} ^{D_{1} \times \frac{H}{4} \times \frac{W}{4}}$, where the total number of patches is always $\frac{H}{4} \times \frac{W}{4} = \frac{HW}{16}$. Finally, the high-resolution features containing both local and global information are utilized for HMR.  

\section{Experiments}
First, we evaluate the effectiveness of our PAT block for the vision transformer backbone, which means we train POTTER without the HR stream for the classification task on ImageNet. The pretrained weights on ImageNet can be used as a good initialization for downstream tasks such as HMR. Next, we train the entire POTTER (with HR stream, as shown in Fig. \ref{fig:main_arch}) for reconstructing human mesh.    

\subsection{Image Classification}
\begin{figure}[htp]
\vspace{-5pt}
  \centering
  \includegraphics[width=1\linewidth]{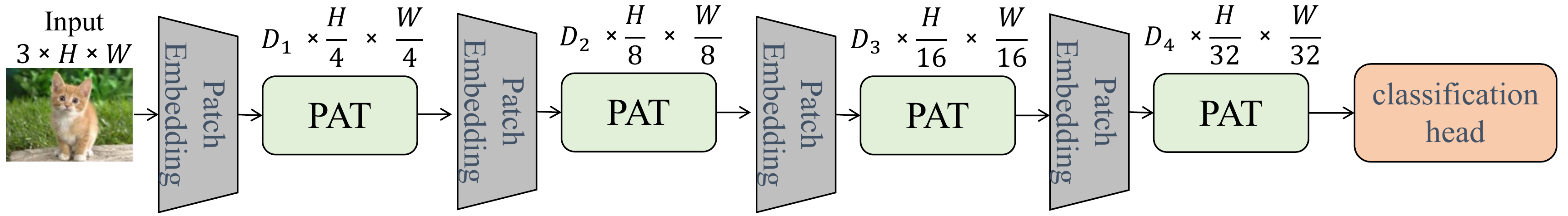}
  \vspace{-15pt}
  \caption{The framework for image classification.}
  \label{fig:cls-framework}
  \vspace{-5pt}
\end{figure}

The framework of POTTER without HR stream for the image classification (named POTTER\_cls) is shown in Fig. \ref{fig:cls-framework} where PAT is utilized as the transformer block.

\textbf{Dataset and Implementation Details:} ImageNet-1k is one of the most commonly used datasets for computer vision tasks, which consists of 1.3M training images and 50K validation images within 1k classes. 

We follow the same training scheme as \cite{deit,poolformer}. Our models are trained for 300 epochs with peak learning rate $lr=2e^{-3}$ and batch size of 1024. AdamW \cite{kingma2014adam} optimizer with cosine learning rate schedule is adopted. We use the same Layer Normalization as implemented in \cite{deit}. More details are provided in the \textcolor{blue}{Supplementary 
 \ref{detail}}.   

\begin{table}[htp]
\vspace{-5pt}
\centering
  \caption{Performance of different types of models on ImageNet-1k classification task.  All these models are only trained on the ImageNet1k training set. The top-1 accuracy on the validation set is reported in this table. More results comparisons are provided in the \textcolor{blue}{Supplementary \ref{detail}}.}
  \resizebox{0.85\linewidth}{!}
  {
\begin{tabular}{c|c|c|c|c}
\hline
               & Image Size & Params (M) & MACs (G) & Top-1 Acc $\uparrow$ \\ \hline
ViT-L/16 \cite{Dosovitskiy2020ViT}       & 224        & 307        & 190.7    & 76.5      \\
RSB-ResNet-18 \cite{RSB-resnet} & 224        & 12         & 1.8      & 70.6      \\
RSB-ResNet-34 \cite{RSB-resnet} & 224        & 22         & 3.7      & 75.5      \\
PVT-Tiny \cite{PVT}      & 224        & 13         & 1.9      & 75.1      \\
MLP-Mixer-B/16 \cite{mlpmixer}    & 224        & 59         & 12.7     & 76.4      \\
ResMLP-S12 \cite{resmlp}    & 224        & 15         & 3.0      & 76.6      \\
PoolFormer-S12 \cite{poolformer} & 224        & 12         & 1.8      & 77.2      \\ \hline
POTTER\_cls           & 224        & 12         & 1.8      & \textbf{79.0}      \\ \hline
\end{tabular}
}
\label{tab: image_cls}
\vspace{-10pt}
\end{table}

\begin{table*}[htp]
\vspace{-10pt}
\centering
  \caption{3D Pose and Mesh performance comparison with SOTA methods on Human3.6M \cite{h36m_pami} and 3DPW \cite{pw3d2018} datasets. 
  The ``*'' indicates that HybrIK \cite{hybrik} uses ResNet34 as the backbone and with predicted camera parameters. The comparison with \cite{Thundr} is provided in the \textcolor{blue}{Supplementary \ref{detail}}.}
  \vspace{-5pt}
  \resizebox{0.92\linewidth}{!}
  {
  \begin{tabular}{c|cc|cc|cc|ccc}
\hline
                                                                              &                &           &            &            & \multicolumn{2}{c|}{Human3.6M} & \multicolumn{3}{c}{3DPW} \\ \cline{2-10} 
                                                                              & Model          & Year      & Params(M)  & MACs(G)    & MPJPE $\downarrow$        & PA-MPJPE $\downarrow$      & MPJPE $\downarrow$ & PA-MPJPE $\downarrow$ & MPVE $\downarrow$ \\ \hline
\multirow{10}{*}{CNN-based}                                                   & HMR \cite{kanazawaHMR18}           & CVPR 2018 & -          & -          & 88.0           & 56.8          & 130.0 & 76.7     & -     \\
                                                                              & GraphCMR \cite{kolotouros2019cmr}          & CVPR 2019 & -          & -          & -              & 50.1          & -     & 70.2     & -     \\
                                                                              & SPIN \cite{Kolotouros2019SPIN}              & ICCV 2019 & -          & -          & 62.5           & 41.1          & 96.9  & 59.2     & 116.4 \\
                                                                              & VIBE \cite{kocabas2020vibe}         & CVPR 2020 & -          & -          & 65.6           & 41.4          & 82.9  & 51.9     & 99.1  \\
                                                                              & I2LMeshNet \cite{Moon_I2L_MeshNet}     & ECCV 2020 & 140.5      & 36.6       & 55.7           & 41.1          & 93.2  & 57.7     & -     \\
                                                                              & HybrIK* \cite{hybrik}         & CVPR2021  & 27.6       & 12.7       & 57.3           & 36.2          & 75.3  & 45.2     & 87.9  \\
                                                                              & ProHMR \cite{prohmr}        & ICCV 2021 & -          & -          & -              & 41.2          & -     & 59.8     & -     \\
                                                                              & PyMAF \cite{pymaf2021}          & ICCV 2021 & 45.2       & 10.6       & 57.7           & 40.5          & 92.8  & 58.9     & 110.1 \\
                                                                              & DSR \cite{dsr2021}           & ICCV 2021 & \textbf{-} & \textbf{-} & 60.9           & 40.3          & 85.7  & 51.7     & 99.5  \\
                                                                              & OCHMR \cite{OCHMR}          & CVPR 2022 & \textbf{-} & \textbf{-} & \textbf{-}     & \textbf{-}    & 89.7  & 58.3     & 107.1 \\ \hline
\multirow{5}{*}{\begin{tabular}[c]{@{}c@{}}Transformer\\ -based\end{tabular}} & METRO \cite{lin2021metro}         & CVPR 2021 & 229.2      & 56.6       & 54.0         & 36.7          & 77.1  & 47.9     & 88.2  \\
                                                                                          & GTRS \cite{gtrs} & ACM MM 2022 & 71.5      & 3.8       & 64.3           & 45.4          & 88.5  & 58.9     & 106.2  \\
                                                                              & TCFormer \cite{tcformer}      & CVPR 2022 & -          & -          & 62.9           & 42.8          & 80.6  & 49.3     & -     \\                                                              & FastMETRO-S  \cite{FastMETRO}    & ECCV 2022 & 32.7       & 8.9          & 57.7           & 39.4          & 79.6  & 49.3     & 91.9  \\
                                                                              & FastMETRO-L \cite{FastMETRO}      & ECCV 2022 & 48.5       & 11.8          & \textbf{53.9}           & 37.3          & 77.9  & 48.3     & 90.6  \\ \cline{2-10} 
                             & POTTER           &           & 16.3  & 7.8  & 56.5      & \textbf{35.1}     & \textbf{75.0}  & \textbf{44.8}     & \textbf{87.4}       \\ \hline
\end{tabular}
}
\label{tab: 3dmesh}
\vspace{-10pt}
\end{table*}
\textbf{Results:} The performance of POTTER\_cls on ImageNet is reported in Table \ref{tab: image_cls}. As a small network focusing on efficiency,  POTTER\_cls achieves superior performance with only 12M parameters and 1.8G MACs compared with the commonly-used CNN and transformer-based models. The CNN model RSB-ResNet-34 \cite{RSB-resnet} (ResNet \cite{resnet} trained with improved training procedure for 300 epochs) only achieves 75.5 \% of the top-1 accuracy. The transformer-based methods with small params and MACs such as PVT-tiny \cite{PVT}, MLP-Mixer-S12 \cite{mlpmixer}, and PoolFormer-S12  \cite{poolformer} obtain the top-1 accuracy around  75.1 \% - 77.2 \%. Our  POTTER\_cls outperforms them by achieving 79.0 \% of top-1 accuracy with fewer Params and MACs. 
PoolFormer \cite{poolformer} demonstrates that using an extremely simple pooling layer without attention design as a patch mixer can still achieve highly competitive performance. We claim that an efficient pooling-based attention mechanism can further boost performance. With our PoolAttn design, POTTER\_cls outperforms PoolFormer-S12 by 1.8\% without increasing the memory and computational costs. \textit{Thus, POTTER has the potential to enhance other tasks, including HMR.}

\subsection{HMR}
After pretaining on the ImageNet with POTTER\_cls, we load the pretrained weight and train the entire network POTTER with the architecture illustrated in Fig. \ref{fig:main_arch} for the HMR task. An HMR head HybrIK \cite{hybrik} is applied for generating the final human mesh.

\textbf{ Dataset and Implementation Details:} 
Human3.6M \cite{h36m_pami} is one of the commonly used datasets with the indoor setting which consists of 3.6M video frames performed by 11 actors. Following \cite{Kolotouros2019SPIN,Moon_I2L_MeshNet,tcformer,lin2021metro}, there are 5 subjects (S1, S5, S6, S7, S8) used for training and 2 subjects (S9, S11) used for testing. 3DPW \cite{pw3d2018} a widely used outdoor dataset that contains 60 video sequences with 51k frames. Unlike Human3.6M which only provides the annotation of key joints, accurate 3D mesh annotation is available for the 3DPW. Thus, we evaluate the Mean Per Joint Position Error (MPJPE) \cite{hmrsurvey} and MPJPE after Procrustes Alignment (PA-MPJPE) \cite{hmrsurvey} on Human3.6M. For 3DPW, we report the MPJPE, PA-MPJPE, and e Mean Per Vertex Error (MPVE). Following previous work \cite{hybrik,FastMETRO,lin2021metro, lin2021_mesh_graphormer}, Human3.6M, MPI-INF-3DHP \cite{mpi3dhp2017}, COCO \cite{lin2014mscoco}, and 3DPW are used for mixed training. We train our POTTER with the architecture illustrated in Fig. \ref{fig:main_arch}. 
The Adam \cite{kingma2014adam} optimizer is utilized for training where the learning rate is $1 \times 10^{-3}$ with a batch size of 32. All experiments are conducted on four NVIDIA RTX5000 GPUs with Pytorch\cite{PyTorch} implementation. The 3D joint loss and vertex loss are applied during the training. More implementation details are provided in the \textcolor{blue}{Supplementary \ref{detail}}.

\textbf{Results:}
Table \ref{tab: 3dmesh} compares POTTER with previous SOTA methods for the HMR task on Human3.6M and 3DPW datasets. As a pure transformer architecture, POTTER outperforms the previous transformer-based SOTA method METRO \cite{lin2021metro} (a hybrid CNN+transformer architecture) by showing significant computational and memory reduction. To be more specific, POTTER only requires 16.3 M Params and 7.8 G MACs (\textbf{7\% of Params and 14\% MACs compared with METRO}) to achieve the new SOTA performance. Although FastMETRO \cite{FastMETRO} reduces the computational and memory costs of METRO, it is still much more expensive than our POTTER with worse performance. \textit{Without bells and whistles, POTTER demonstrates its exceptional performance both in terms of mesh recovery accuracy and model efficiency. }

\begin{figure}[htp]
\vspace{-5pt}
  \centering
  \includegraphics[width=0.9\linewidth]{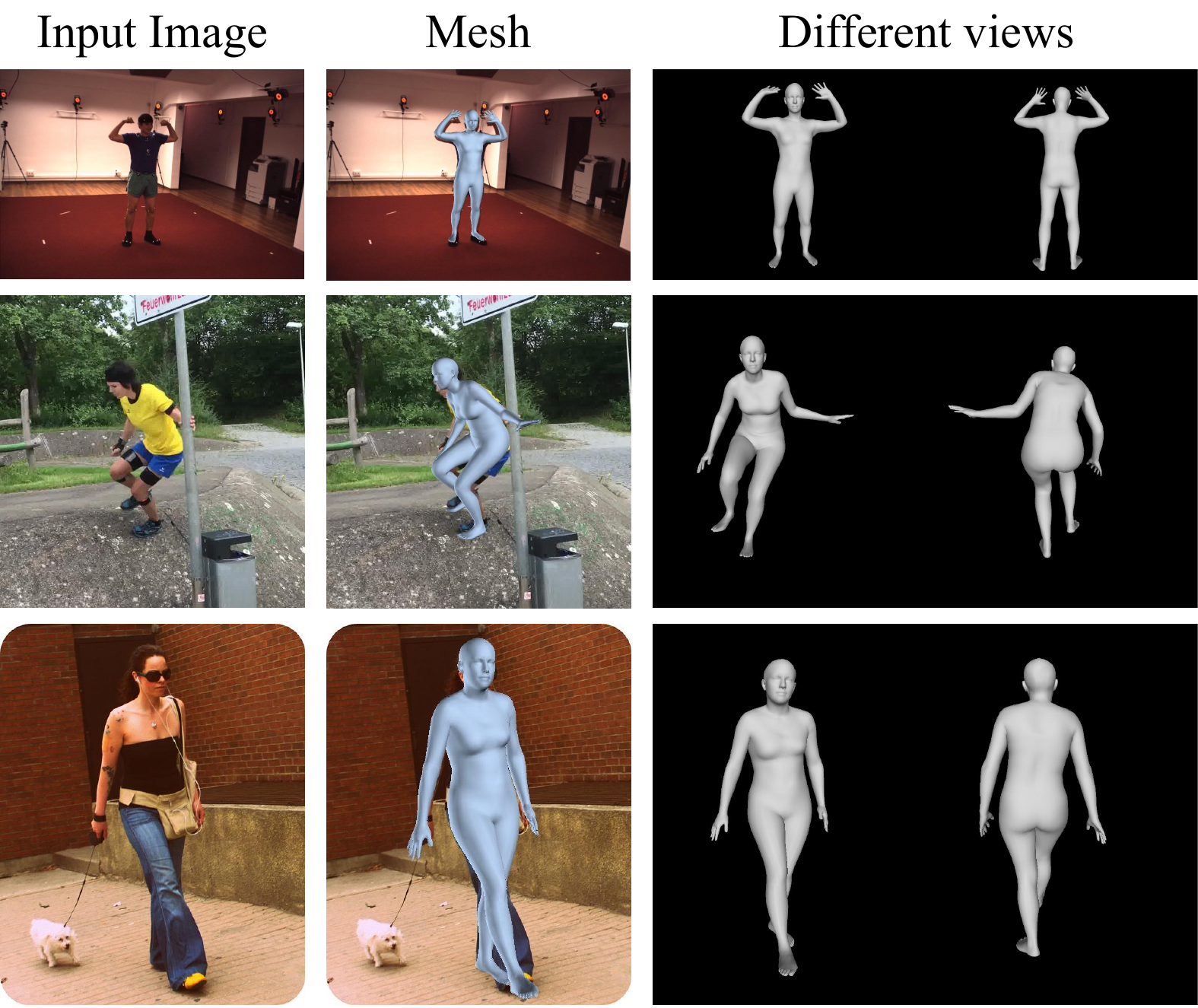}
  \vspace{-10pt}
  \caption{Mesh visualizations of POTTER. Images are taken from Human3.6M, 3DPW, and COCO datasets.}
  \label{fig:vis_main}
  \vspace{-10pt}
\end{figure}

\textbf{Mesh visualization:}
We show qualitative results of POTTER on Human3.6M, 3DPW, and COCO dataset in Fig. \ref{fig:vis_main}. POTTER can estimate reliable human poses and meshes given various input images. More qualitative results are presented in the \textcolor{blue}{Supplementary \ref{meshvis}}.

\subsection{Ablation Study}
\label{sec:Ablation}
\textbf{Effectiveness of Pooling Attention Design}: First, we verify the effectiveness of the PoolAttn module proposed in Section \ref{sec:PoolAttnDesign}. 
We report the top-1 accuracy using the different block combinations on the ImageNet classification task. ``Pooling" denotes that only a pooling layer is used without any pooling attention design (which is the exact architecture of the PoolFormer \cite{poolformer}). The top-1 accuracy is 77.2 \% with 11.9 M Params and 1.79 G MACs. When only applying patch-wise pooling attention, the performance is improved by 1.5 \% with almost the same Params and MACs.  Similarly, when only applying embed-wise pooling attention, the performance is also improved by 1.3 \% with a slight increase in the Params and MACs. For our PoolAttn module which integrates two types of pooling attention, the performance is further boosted by 1.8 \%, which demonstrates the effectiveness of the PoolAttn in the backbone. 

\begin{table}[htp]
\vspace{-5pt}
\centering
  \caption{Ablation study of different modules in the transformer block on ImageNet classification. }
  \vspace{-10pt}
  \resizebox{0.9\linewidth}{!}
  {
\begin{tabular}{c|ccc}
\hline
Module                                                                                            & Params(M) & MACs(G) & Top-1 Acc $\uparrow$ \\ \hline
Pooling                                                                                          & 11.9      & 1.79    & 77.2      \\ \hline
Patch-Wise Pooling Attention                                                                     & 12.0      & 1.82    &  78.7          \\ \hline
Embed-Wise Pooling Attention                                                                     & 12.2      & 1.83    &  78.5         \\ \hline
PoolAttn (Patch-Wise and Embed-Wise)  & 12.4      & 1.84    & 79.0      \\ \hline

\end{tabular}
}
\label{tab: ab_block}
\vspace{-5pt}
\end{table}

\begin{table}[htp]
\vspace{-5pt}
\centering
  \caption{Ablation study of different modules in the transformer block on 3DPW dataset for HMR.}
  \vspace{-10pt}
  \resizebox{0.9\linewidth}{!}
  {
\begin{tabular}{c|cc|ccc}
\hline
         &           &         & \multicolumn{3}{c}{3DPW} \\ \hline
Module    & Params(M) & MACs(G) & MPJPE $\downarrow$  & PA-MPJPE $\downarrow$ & MPVE $\downarrow$ \\ \hline
Pooling  & 15.9      & 7.7     & 77.3   & 47.4     & 89.9 \\ \hline
PoolAttn & 16.3      & 7.8     & 75.0   & 44.8     & 87.4 \\ \hline
\end{tabular}
}
\label{tab: ab_poolattn_hmr}
\vspace{-5pt}
\end{table}

Next, we evaluate the effectiveness of the PoolAttn on the HMR task. As shown in Table \ref{tab: ab_poolattn_hmr}, replacing the pooling block (PoolFormer's architecture) by our PoolAttn, the performance of all metrics (MPJPE, PA-MPJPE, and MPVE) on 3DPW dataset can be improved. The increase in memory and computational cost can be neglected. Thus, PoolAttn design is critical for efficient HMR with high accuracy. 

\begin{figure}[htp]
\vspace{-5pt}
  \centering
  \includegraphics[width=0.8\linewidth]{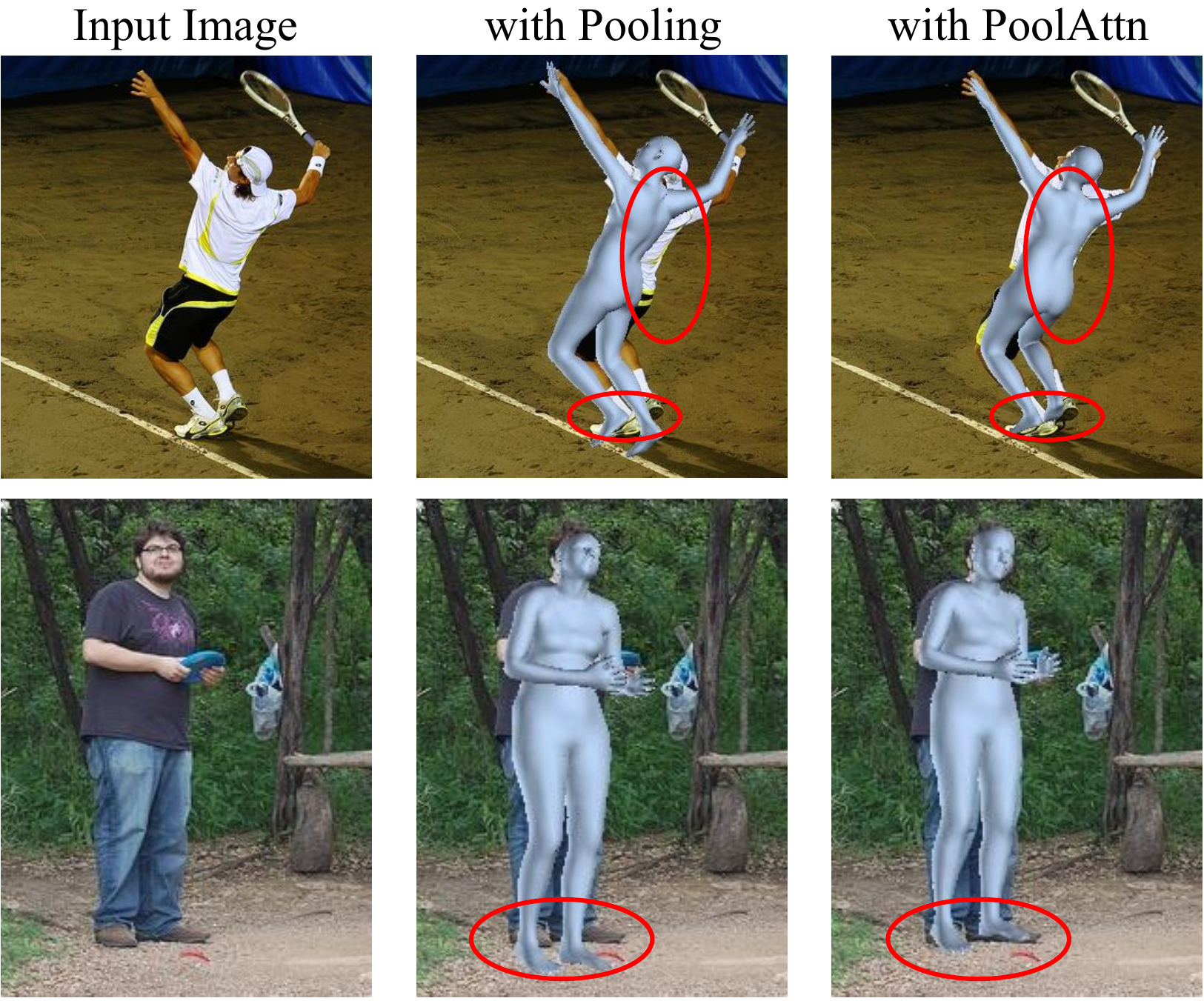}
  \vspace{-5pt}
  \caption{The visual comparison between applying the Pooling layer and PoolAttn layer. The red circles highlight locations where PoolAttn is more accurate than Pooling.}
  \label{fig:vis_poolattn}
  \vspace{-5pt}
\end{figure}

We also show the visualization of using the Pooling layer compared with using PoolAttn layer in Fig. \ref{fig:vis_poolattn}. The areas highlighted by red circles indicate that PoolAttn outputs more accurate meshes than Pooling.

\textbf{Effectiveness of HR stream}: 
We investigate the use of the HR stream in Table \ref{tab: ab_hr}. If we use the Pooling block in transformer (PoolFormer's architecture \cite{poolformer}), the results can be improved by a large margin (3.8 of MPJPE, 2.1 of PA-MPJPE, and 3.6 of MPVE) when adding the HR stream. If we apply our PoolAttn module in transformer (our PAT design), the HR stream can also boost the performance notably (2.8 of MPJPE, 0.8 of PA-MPJPE, and 2.4 of MPVE). 

\begin{figure}[htp]
\vspace{-5pt}
  \centering
  \includegraphics[width=0.9\linewidth]{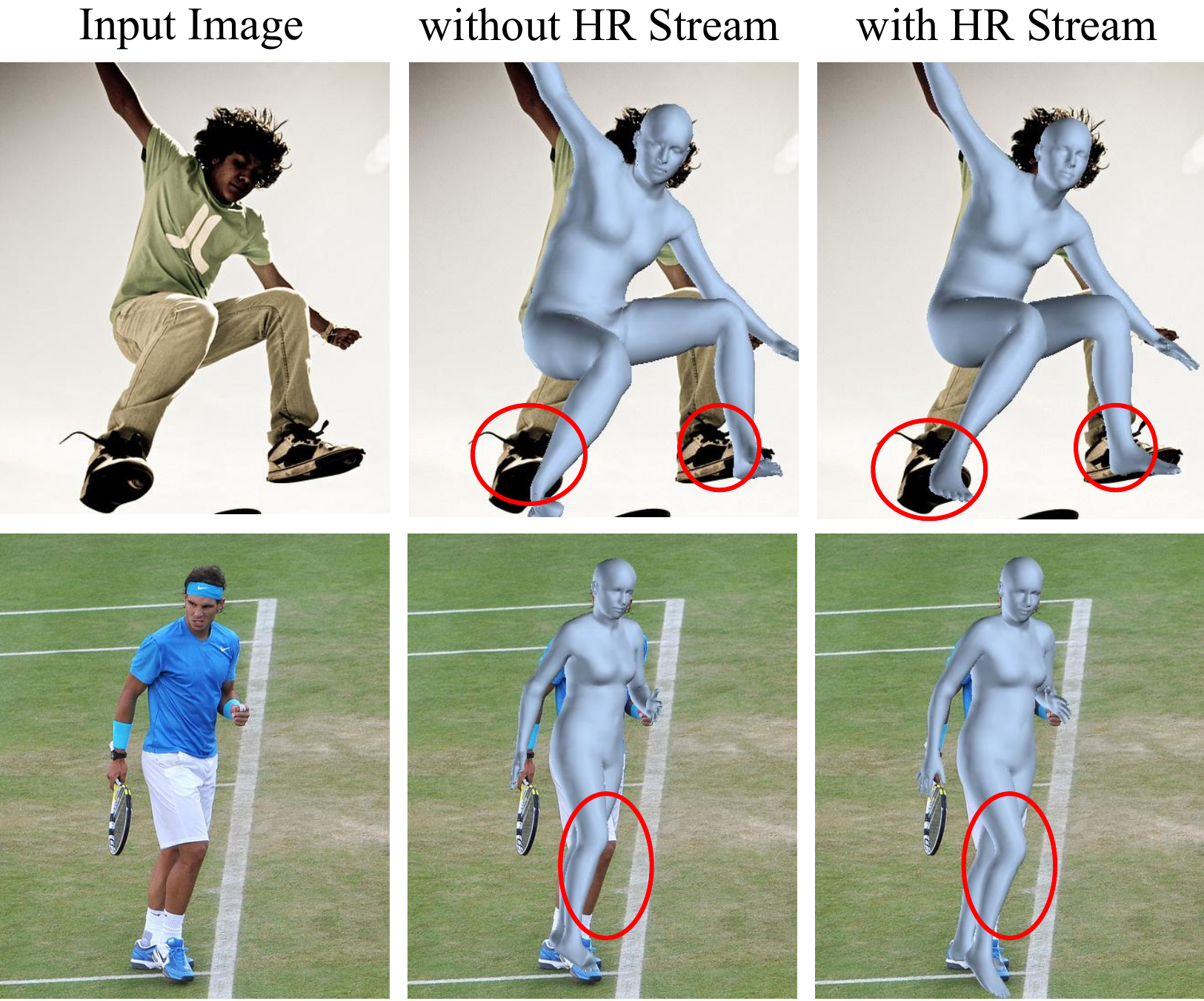}
  \vspace{-5pt}
  \caption{The visual comparison between with HR stream and without HR stream. The red circles highlight locations where POTTER with HR stream is more accurate. }
  \label{fig:vis_hr}
  \vspace{-7pt}
\end{figure}

We further compare the mesh visualization of POTTER (with HR stream) and without HR stream as shown in Fig. \ref{fig:vis_hr}. The areas highlighted by red circles indicate that the HR stream can improve the quality of the reconstructed mesh.

\begin{table}[htp]
\vspace{-5pt}
\centering
  \caption{Ablation study of the HR stream in the transformer architecture on 3DPW dataset for HMR.}
  \vspace{-5pt}
  \resizebox{0.8\linewidth}{!}
  {
\begin{tabular}{cc|ccc}
\hline
                                               &             & \multicolumn{3}{c}{3DPW} \\ \hline
\multicolumn{1}{c|}{Block}                     & Architecture & MPJPE $\downarrow$     & PA-MPJPE $\downarrow$       & MPVE $\downarrow$      \\ \hline
\multicolumn{1}{c|}{\multirow{2}{*}{Pooling}}  & Without HR-stream        & 81.1          &   49.5          &   93.5         \\ \cline{2-5} 
\multicolumn{1}{c|}{}                          & With HR-stream    &  77.3         &   47.4          &   89.9         \\ \hline
\multicolumn{1}{c|}{\multirow{2}{*}{PoolAttn}} & Without HR-stream        &  77.8         &   45.6          &   89.8         \\ \cline{2-5} 
\multicolumn{1}{c|}{}                          & With HR-stream    & 75.0          &  44.8           &   87.4         \\ \hline
\end{tabular}
}
\label{tab: ab_hr}
\vspace{-10pt}
\end{table}

\section{Conclusion}
In this paper, we present POTTER, a pure transformer-based architecture for efficient HMR. A Pooling Attention Transformer block is proposed to reduce the memory and computational cost without sacrificing performance. Moreover, an HR stream in the transformer architecture is proposed to return high-resolution local and global features for the HMR task. Extensive experiments show that POTTER achieves SOTA performance on challenging HMR datasets while significantly reducing the computational cost. 

As an image-based method, one limitation is that POTTER can not fully exploit the temporal information given video sequences input. We will extend POTTER to a video-based version that can output smooth 3D motion with better temporal consistency.  

\textbf{Acknowledgement}: This work is supported by gift funding from OPPO Research.

\newpage
\noindent \textbf{\large{Supplementary Material}}

In this supplementary material, we provide the following sections:

\begin{itemize}

\item Section \ref{Broader}: Broader Impact and Limitations.

\item Section \ref{meshvis}: Human Mesh Visualization on in-the-wild data.

\item Section \ref{supp_Complexity}: Memory and Computational Costs of One PAT Block.

\item Section \ref{detail}: More Experiments (image classification and HMR) and Implementation Details.

\item Section \ref{hand}: Generalization to 3D Hand Reconstruction

\end{itemize}

\appendix




\section{Broader Impact and Limitations}
\label{Broader}
We anticipate that our POTTER can be used for widespread applications such as motion capture in animation and movies, virtual AI assistants, and VR/AR content. Currently, motion capture devices are mandatory for these applications, which are usually expensive, time-consuming, and complicated to set up. In contrast, one of the biggest advantages of our method is that POTTER can reconstruct 3D human mesh directly from 2D images/videos without extra devices. With the reliable reconstructing quality as depicted in Section \ref{meshvis}, POTTER shows a promising impact as a lightweight model for real-world applications. 

There are also a few limitations of POTTER. Although POTTER can estimate reliable human mesh for in-the-wild scenarios, the performance would be downgraded when a severe occlusion exists. Another challenge is POTTER may fail for the rare and complicated pose scenarios due to limited training data. We will tackle these issues in future work.

\begin{figure*}[htb]
\vspace{-5pt}
  \centering
  \includegraphics[width=0.95\linewidth]{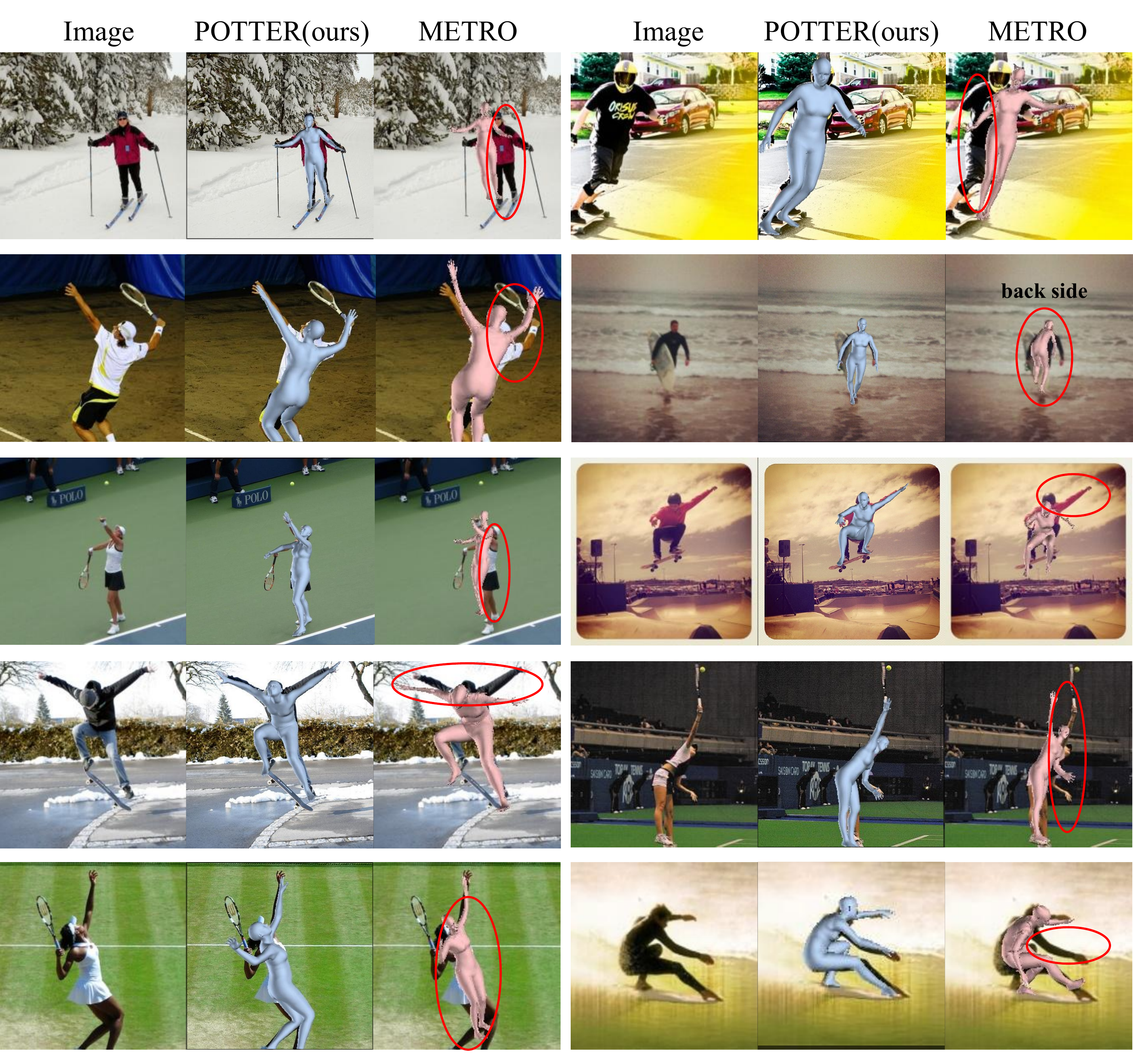}
  \vspace{-10pt}
  \caption{Qualitative comparison with SOTA transformer-based method METRO \cite{lin2021metro}. The \textcolor{red}{red circles} highlight regions where our POTTER generates more accurate mesh recoveries than METRO. Images are taken from the in-the-wild COCO \cite{lin2014mscoco} dataset. }
  \label{fig:supp_comp}
  \vspace{-15pt}
\end{figure*}

\begin{figure*}[htb]
\vspace{-5pt}
  \centering
  \includegraphics[width=0.95\linewidth]{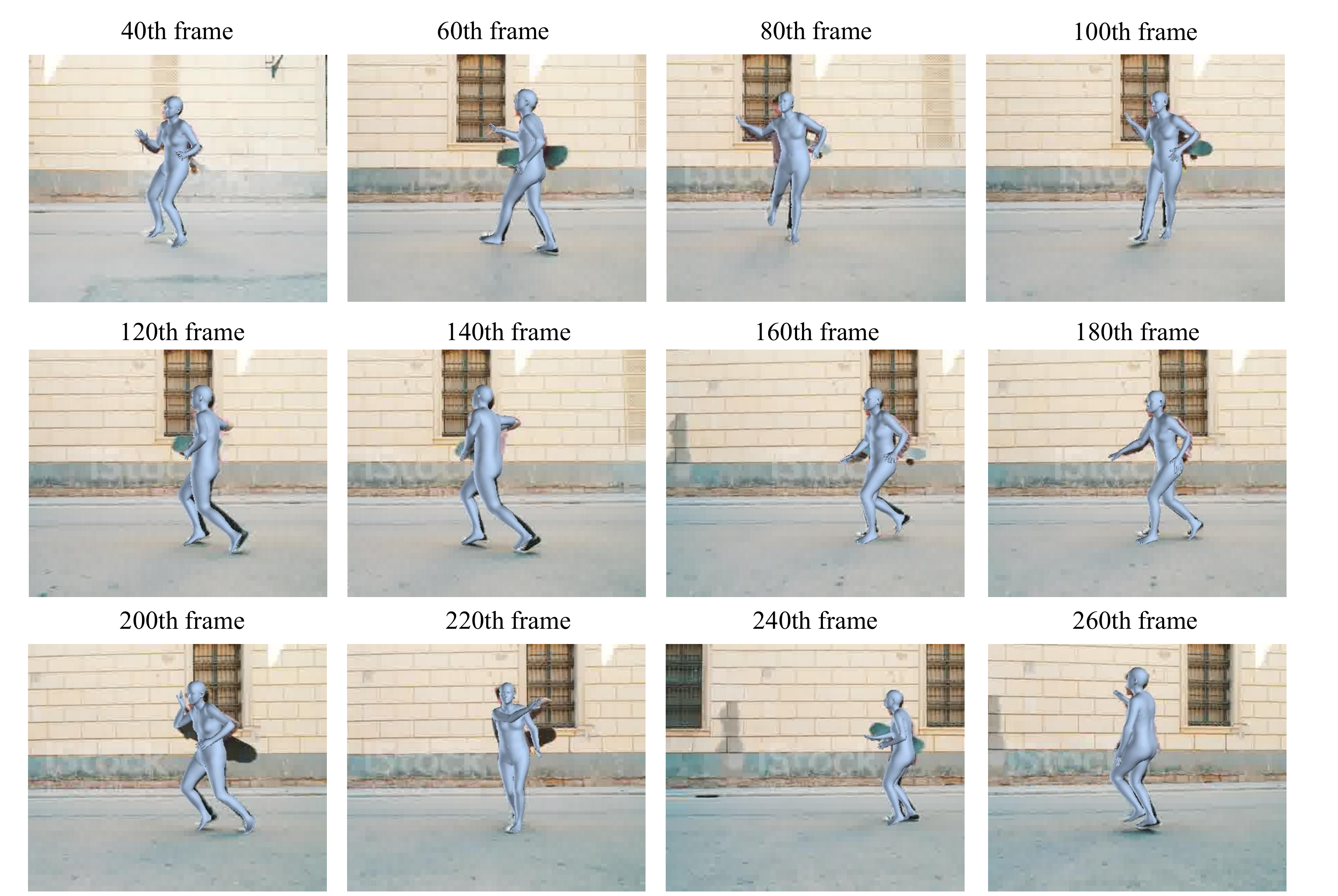}
  \caption{Qualitative results of using POTTER to reconstruct human mesh from an in-the-wild video. Although POTTER is an image-based method, the frame-by-frame reconstruction still works well. \textcolor{blue}{Please refer to our \textbf{video demo} for the reconstructed mesh sequences.} }
  \label{fig:supp_video}
  \vspace{-5pt}
\end{figure*}

\section{Human Mesh Visualization on in-the-wild data}
\label{meshvis}
POTTER achieves superior performance on Human3.6M \cite{h36m_pami} and 3DPW \cite{pw3d2018} datasets as described in the main paper. However, it is critical to evaluate the actual performance of our POTTER on in-the-wild data. Reconstructing accurate human mesh on in-the-wild data is an extremely challenging task due to the different human shapes, scales, pose variations, and backgrounds from the training data. 

In Fig. \ref{fig:supp_comp}, we show the qualitative comparison with SOTA transformer-based method METRO \cite{lin2021metro} in this challenging scenarios (images are taken from in-the-wild dataset COCO \cite{lin2014mscoco}). Following METRO, we use the SMPL gender-neutral model \cite{SMPL:2015} for all visualization. Our POTTER clearly outperforms METRO in many challenging cases, where the red circles highlight the area where POTTER is more accurate than METRO.  

As an image-based method, POTTER can also reconstruct human mesh sequences given the input videos.  In Fig. \ref{fig:supp_video}, we select several frames of the reconstructed human mesh to illustrate the performance of POTTER. We also provide the \textbf{video demo} of the entire reconstructed sequence in the supplementary material, which demonstrates the effectiveness of POTTER given the \textbf{in-the-wild} videos. 

\begin{figure}[htp]
\vspace{-5pt}
  \centering
  \includegraphics[width=0.95\linewidth]{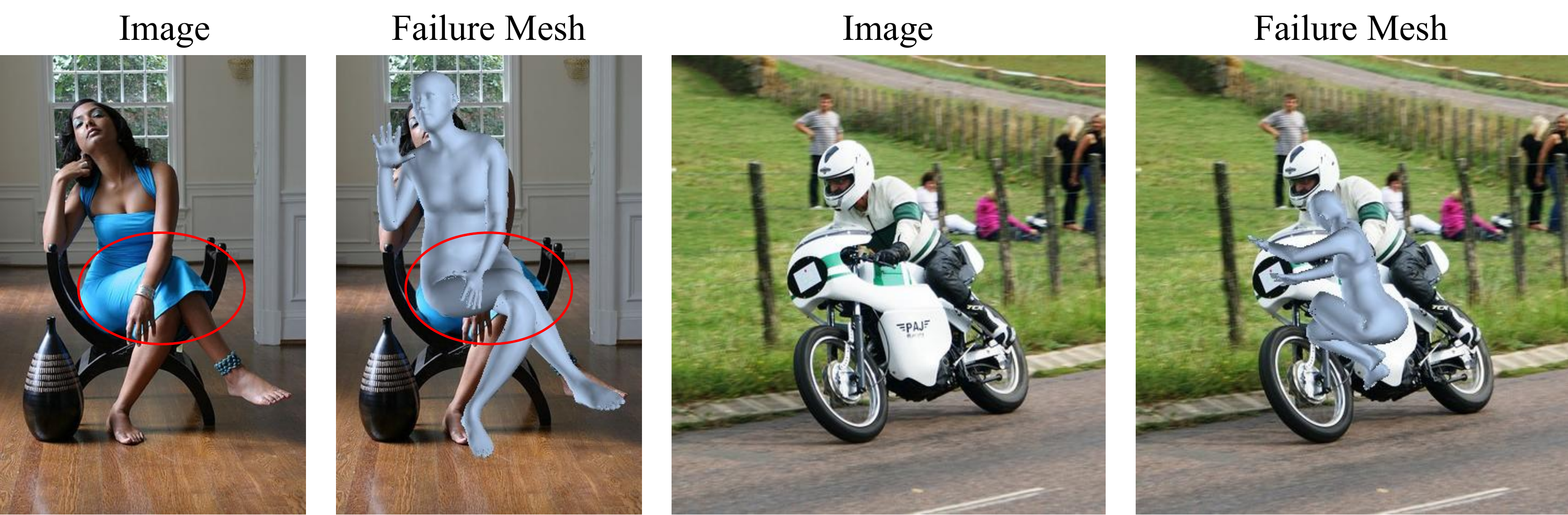}
  \caption{Failure cases. POTTER may not perform well due to severe occlusion. }
  \label{fig:supp_fail}
  \vspace{-5pt}
\end{figure}

Since POTTER is a data-driven approach, the performance can not be guaranteed if the image is very different from the training data (i.e. data distribution shift), such as complicated pose and heavy occlusion (see Fig. \ref{fig:supp_fail} for an example). How to tackle these issues would be our future work. One potential solution is to use the domain adaptation method to make the trained model adapt to the target domain for better mesh recovery.

\begin{table}[htp]
\scriptsize
\renewcommand\arraystretch{1.4}
\centering
  \caption{Total parameters and MACs of one PAT block. }
  \vspace{-10pt}
  \resizebox{0.9\linewidth}{!}
  {
\begin{tabular}{cccc|c|c}
\hline
\multicolumn{4}{c|}{Layer}                                                                                                                                & Params                & MACs                   \\ \hline
\multicolumn{1}{c|}{\multirow{11}{*}{PAT}} & \multicolumn{1}{c|}{\multirow{9}{*}{PoolAttn}} & \multicolumn{1}{c|}{\multirow{4}{*}{Patch-wise}} & Pooling1 & \multirow{4}{*}{$10D$}  & \multirow{4}{*}{$9DN$}   \\
\multicolumn{1}{c|}{}                      & \multicolumn{1}{c|}{}                          & \multicolumn{1}{c|}{}                            & Pooling2 &                       &                        \\
\multicolumn{1}{c|}{}                      & \multicolumn{1}{c|}{}                          & \multicolumn{1}{c|}{}                            & MatMul   &                       &                        \\
\multicolumn{1}{c|}{}                      & \multicolumn{1}{c|}{}                          & \multicolumn{1}{c|}{}                            & Proj1    &                       &                        \\ \cline{3-6} 
\multicolumn{1}{c|}{}                      & \multicolumn{1}{c|}{}                          & \multicolumn{1}{c|}{\multirow{4}{*}{Embed-wise}} & Pooling3 & \multirow{4}{*}{$10D$}  & \multirow{4}{*}{$9DN$}   \\
\multicolumn{1}{c|}{}                      & \multicolumn{1}{c|}{}                          & \multicolumn{1}{c|}{}                            & Pooling4 &                       &                        \\
\multicolumn{1}{c|}{}                      & \multicolumn{1}{c|}{}                          & \multicolumn{1}{c|}{}                            & MatMul   &                       &                        \\
\multicolumn{1}{c|}{}                      & \multicolumn{1}{c|}{}                          & \multicolumn{1}{c|}{}                            & Proj2    &                       &                        \\ \cline{3-6} 
\multicolumn{1}{c|}{}                      & \multicolumn{1}{c|}{}                          & \multicolumn{1}{c|}{Projection}                  & Proj3    & $10D$                   & $9DN$                    \\ \cline{2-6} 
\multicolumn{1}{c|}{}                      & \multicolumn{2}{c|}{\multirow{2}{*}{FFN}}                                                         & MLP1     & $4D^2$ & $4D^2 N$ \\ \cline{4-6} 
\multicolumn{1}{c|}{}                      & \multicolumn{2}{c|}{}                                                                             & MLP2      & $4D^2$ & $4D^2 N$ \\ \hline
\end{tabular}
}
\label{tab: appd_complexity}
\vspace{-10pt}
\end{table}
\section{Memory and Computational Costs of One PAT Block}
\label{supp_Complexity}
To achieve model efficiency, one PAT block in the proposed method  consists of one PoolAttn module with a Feedforward Network (FFN). For the layers such as pooling, layer normalization, and matrix multiplication operations for the squeezed features, the required memory and computational costs can be ignored when compared with the projection or FFN layer. Thus, The total parameters and MACs of one PAT block given the input $[D,h,w]$ can be estimated as in Table \ref{tab: appd_complexity}, where the number of patches $N=h \times w$. To save the memory and computational costs, we utilize depth-wise convolution \cite{chollet2017xception} served as the ``Proj1", ``Proj2", and ``Proj3". The PoolAttn only requires $10D$ params and $9DN$ MACs. Compared with the conventional attention module which requires $(4D^2 + 4D)$ params and $(4DN^2+2D^2N)$ MACs, our PoolAttn significantly reduce the complexity from   $\mathcal{O}(D^2)$ to $\mathcal{O}(D)$.

\begin{figure*}[htp]
\vspace{-15pt}
  \centering
  \includegraphics[width=0.98\linewidth]{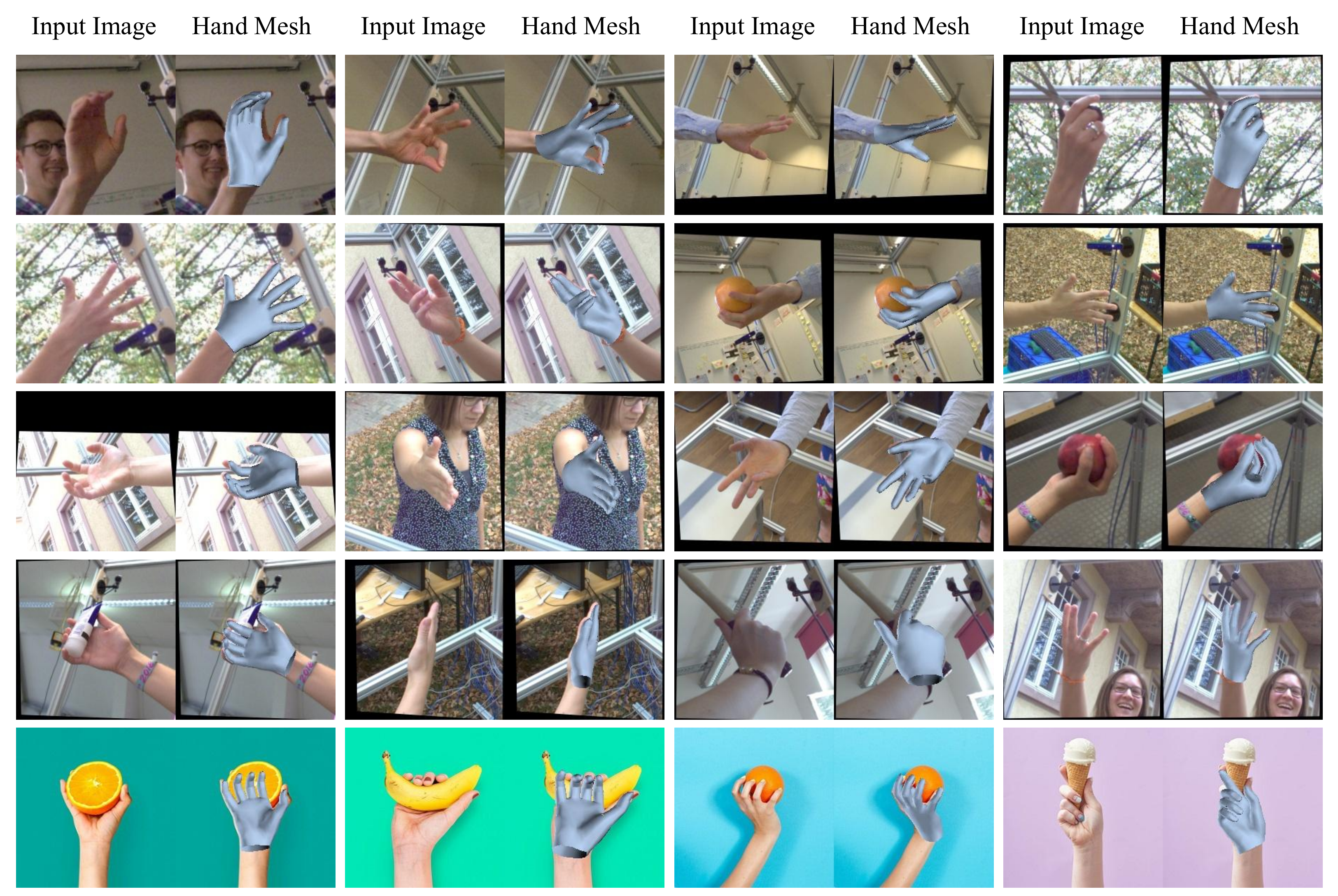}
  \vspace{-10pt}
  \caption{Qualitative results of our POTTER for reconstructing hand mesh.}
  \label{fig:supp_hand}
  \vspace{-15pt}
\end{figure*}


\section{More Experiments and Implementation Details}
\label{detail}
\subsection{Image Classification}
For the image classification task, we follow the same training scheme as PoolFormer \cite{poolformer}.  Our model POTTER\_{cls} is trained for 300 epochs with a cosine learning rate schedule (The number of warm-up epochs is 5). The AdamW optimizer \cite{kingma2014adam} is used with weight decay 0.05 and peak learning rate $lr=1e^{-3}$ and batch size 1024. The input image is with the size of [224, 224]. For POTTER\_{cls}, the number of blocks for each stage is [2,2,6,2], which is the same as PoolFormer-S12. POTTER\_{cls} outperforms PoolFormer-S12 by 1.8 \% without increasing the memory and computational costs. 

To further verify that our pooling attention design can significantly reduce the memory and computational cost without sacrificing performance, we increase the number of blocks for each stage as [4,4,12,4], named POTTER\_{cls}\_S24. The result is shown in Table \ref{tab: supp_image_cls}. With the same hierarchical architecture, POTTER\_{cls}\_S24 (with PoolAttn) surpasses Swin-Tiny (with conventional attention) by requiring 72\% of Params and 78\% of MACs.

\begin{table}[htp]
\vspace{-5pt}
\centering
  \caption{Performance of different types of models on ImageNet-1K classification task.  All these models are only trained on the ImageNet1K training set. The top-1 accuracy on the validation set is reported in this table.}
  \resizebox{0.95\linewidth}{!}
  {
\begin{tabular}{c|c|c|c|c}
\hline
               & Image Size & Params (M) & MACs (G) & Top-1 Acc $\uparrow$ \\ \hline
RSB-ResNet-50 \cite{RSB-resnet} & 224        & 26         & 4.1      & 79.8      \\
DeiT-S \cite{RSB-resnet} & 224        & 22         & 4.6      & 79.8      \\
MLP-Mixer-B/16 \cite{mlpmixer}    & 224        & 59         & 12.7     & 76.4      \\
PVT-Small \cite{PVT}      & 224        & 25         & 3.8      & 79.8      \\
ResMLP-S24 \cite{poolformer} & 224        & 30         & 6.0     & 79.4      \\
PoolFormer-S24 \cite{poolformer} & 224        & 21         & 3.4      & 80.3      \\ \hline
Swin-Mixer-T/D6 \cite{liu2021Swin}    & 224        & 23         & 4.0      & 79.7      \\
Swin-Tiny \cite{liu2021Swin} & 224        & 29         & 4.5      & 81.3      \\ \hline
POTTER\_cls\_S24           & 224        & 21         & 3.5      & \textbf{81.4}      \\ \hline
\end{tabular}
}
\label{tab: supp_image_cls}
\vspace{-10pt}
\end{table}

\subsection{Human Mesh Recovery}
For HMR task, the SMPL model \cite{SMPL:2015} is utilized for reconstructing human mesh. Given the predicted pose parameters $\theta$ and the shape parameters $\beta$, the SMPL model can return the body mesh $M \in  \mathbb{R}^{N \times 3}  $ with $N=6890$ vertices by the function $M = SMPL (\theta, \beta) $. After obtaining the body mesh $M$, the body joints $J$ can be regressed by the predefined joint regression matrix $W$, which means   $J \in  \mathbb{R}^{k \times 3} = W \cdot M  $, where $k$ is the number of joints.
The overall loss during the HMR task can be defined as: 

\begin{equation}
\begin{aligned} 
  \begin{aligned}
\mathcal{L}_{HMR} &=  w_1 \| \beta - \beta^*  \| + w_2 \| \theta - \theta^*  \| + w_3 \| J - J^*  \|\\
  \end{aligned}\\
\end{aligned}
\end{equation}

\noindent where * denote the ground-truth value. In our experiments, we set $w_1=0.01$, $w_2=0.01$, and $w_3=1$.

\begin{figure}[htp]
\vspace{-5pt}
  \centering
  \includegraphics[width=0.98\linewidth]{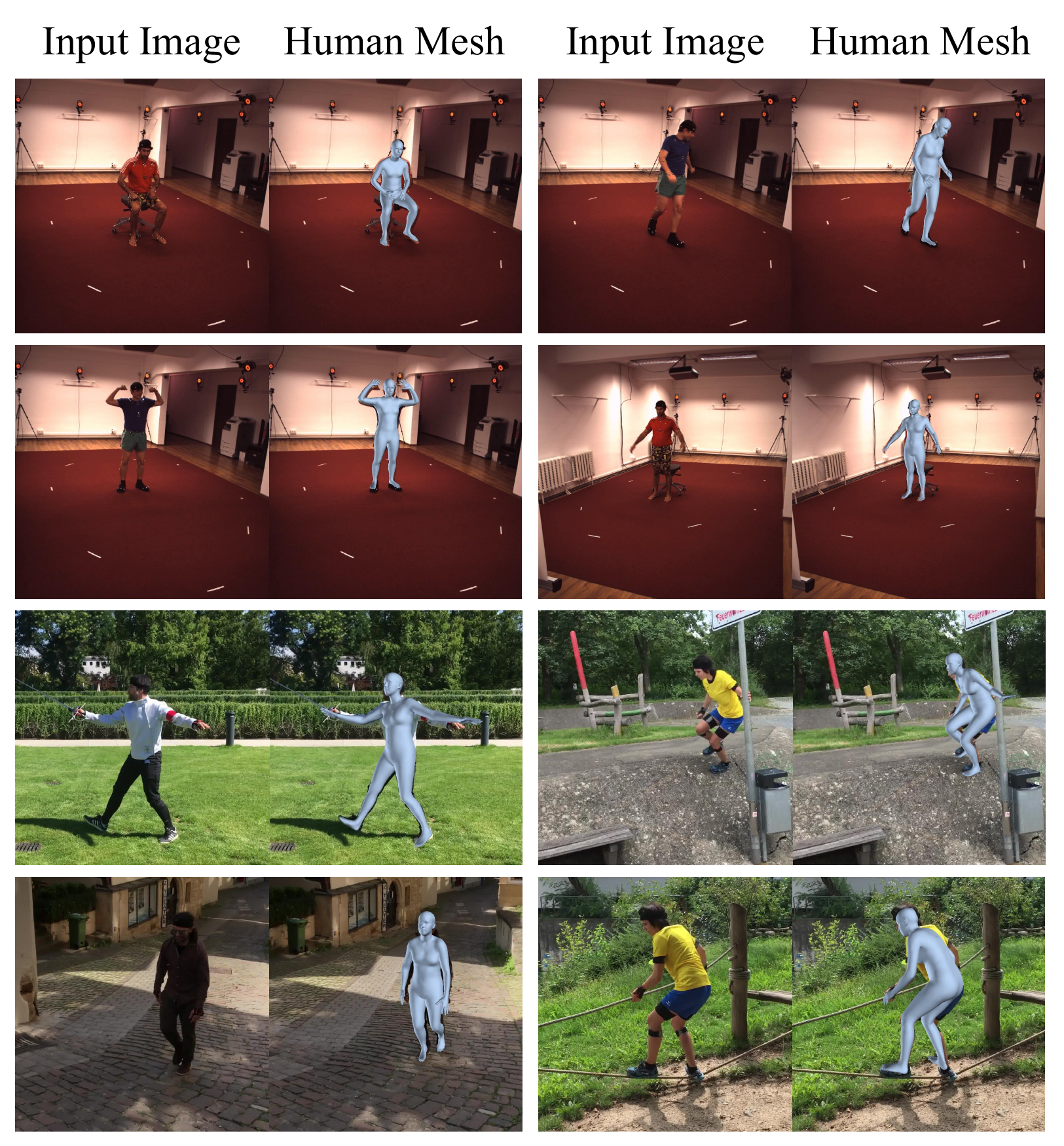}
  \vspace{-10pt}
  \caption{More qualitative results of our POTTER for HMR. Images are taken from Human3.6M and 3DPW datasets}
  \label{fig:supp_3dpw}
  \vspace{-5pt}
\end{figure}

Our POTTER  is trained for 80 epochs with a step learning rate schedule with $lr = 5e-4$ and  $lr_{decay} = 0.1$. The Adam \cite{kingma2014adam} optimizer is utilized for training and the batch size is 32.  The input image is resized to $256 \times 256$. We show more qualitative results for POTTER on images from Human3.6M and 3DPW datasets in Fig. \ref{fig:supp_3dpw}.

Specifically, we compare our POTTER with THUNDR \cite{Thundr} in Table \ref{tab: ab_th}. Since the code of THUNDR is not released, we are unable to compute the MACs. POTTER achieves on-par results compared with THUNDR with 65 \% of total parameters as shown in table \ref{tab: ab_th}. We also notice that THUNDR uses the more recent GHUM Model for the human mesh regression, while our POTTER and other methods such as SPIN \cite{Kolotouros2019SPIN}, DSR \cite{dsr2021}, and TCFormer \cite{tcformer} use the SMPL Model for human mesh regression. This might be the reason that THUNDR achieves better performance. 

\begin{table}[htp]
\centering
  \caption{3D Pose and Mesh performance comparison with SOTA methods on Human3.6M and 3DPW datasets. }
  \resizebox{1\linewidth}{!}
  {
\begin{tabular}{c|cc|cc|ccc}
\hline
       &            &          & \multicolumn{2}{c|}{Human3.6M} & \multicolumn{3}{c}{3DPW}                      \\ \hline
       & Params (M) & MACs (G) & MPJPE          & PA-MPJPE      & MPJPE         & PA-MPJPE      & MPVE          \\ \hline
METRO  & 229.2      & 56.6     & \underline{54.0}     & 36.7          & 77.1          & \underline{47.9}    & 88.2          \\ \hline
THUNDR & 25         & -        & \textbf{48.0}  & \textbf{34.9} & \textbf{74.8} & 51.5          & \underline{88.0}    \\ \hline
POTTER & 16.3       & 7.8      & 56.5           & \underline{35.1}    & \underline{75.0}    & \textbf{44.8} & \textbf{87.4} \\ \hline
\end{tabular}
}

\label{tab: ab_th}
\vspace{-5pt}
\end{table}

\section{Generalization to 3D Hand Reconstruction}
\label{hand}
POTTER can be also generalized for other mesh reconstruction tasks such as 3D hand reconstruction. To demonstrate this capability, we conduct the experiment on the hand mesh dataset FreiHand \cite{freihand}. Without involving extra training data, POTTER can reconstruct reliable hand mesh. Unfortunately, due to the FreiHand online evaluation server being closed (The CodaLab website announced that the server is no longer accepting new challenges not new submissions to old challenges), we are not able to report the test results. Here we provide the hand mesh visualization of POTTER in Fig. \ref{fig:supp_hand}, which demonstrate that  POTTER can generalize well for other tasks such as hand mesh reconstruction. 

{\small
\bibliographystyle{ieee_fullname}
\bibliography{egbib}

\begin{thebibliography}{10}\itemsep=-1pt

\bibitem{carion2020end}
Nicolas Carion, Francisco Massa, Gabriel Synnaeve, Nicolas Usunier, Alexander
  Kirillov, and Sergey Zagoruyko.
\newblock End-to-end object detection with transformers.
\newblock In {\em European Conference on Computer Vision}, pages 213--229.
  Springer, 2020.

\bibitem{maskformer}
Bowen Cheng, Alex Schwing, and Alexander Kirillov.
\newblock Per-pixel classification is not all you need for semantic
  segmentation.
\newblock {\em Advances in Neural Information Processing Systems},
  34:17864--17875, 2021.

\bibitem{FastMETRO}
Junhyeong Cho, Kim Youwang, and Tae-Hyun Oh.
\newblock Cross-attention of disentangled modalities for 3d human mesh recovery
  with transformers.
\newblock In {\em European Conference on Computer Vision (ECCV)}, 2022.

\bibitem{chollet2017xception}
Fran{\c{c}}ois Chollet.
\newblock Xception: Deep learning with depthwise separable convolutions.
\newblock In {\em Proceedings of the IEEE conference on computer vision and
  pattern recognition}, pages 1251--1258, 2017.

\bibitem{Dosovitskiy2020ViT}
Alexey Dosovitskiy, Lucas Beyer, Alexander Kolesnikov, Dirk Weissenborn,
  Xiaohua Zhai, Thomas Unterthiner, Mostafa Dehghani, Matthias Minderer, Georg
  Heigold, Sylvain Gelly, Jakob Uszkoreit, and Neil Houlsby.
\newblock An image is worth 16x16 words: Transformers for image recognition at
  scale.
\newblock {\em ICLR}, 2021.

\bibitem{dsr2021}
Sai~Kumar Dwivedi, Nikos Athanasiou, Muhammed Kocabas, and Michael~J Black.
\newblock Learning to regress bodies from images using differentiable semantic
  rendering.
\newblock In {\em Proceedings of the IEEE/CVF International Conference on
  Computer Vision}, pages 11250--11259, 2021.

\bibitem{resnet}
Kaiming He, Xiangyu Zhang, Shaoqing Ren, and Jian Sun.
\newblock Deep residual learning for image recognition.
\newblock In {\em Proceedings of the IEEE conference on computer vision and
  pattern recognition}, pages 770--778, 2016.

\bibitem{h36m_pami}
Catalin Ionescu, Dragos Papava, Vlad Olaru, and Cristian Sminchisescu.
\newblock Human3.6m: Large scale datasets and predictive methods for 3d human
  sensing in natural environments.
\newblock {\em IEEE Transactions on Pattern Analysis and Machine Intelligence},
  36(7):1325--1339, jul 2014.

\bibitem{Jiang_2020_CVPR}
Wen Jiang, Nikos Kolotouros, Georgios Pavlakos, Xiaowei Zhou, and Kostas
  Daniilidis.
\newblock Coherent reconstruction of multiple humans from a single image.
\newblock In {\em CVPR}, 2020.

\bibitem{kanazawaHMR18}
Angjoo Kanazawa, Michael~J. Black, David~W. Jacobs, and Jitendra Malik.
\newblock End-to-end recovery of human shape and pose.
\newblock In {\em CVPR}, 2018.

\bibitem{OCHMR}
Rawal Khirodkar, Shashank Tripathi, and Kris Kitani.
\newblock Occluded human mesh recovery.
\newblock {\em Proceedings of the IEEE/CVF Conference on Computer Vision and
  Pattern Recognition (CVPR)}, 2022.

\bibitem{kingma2014adam}
Diederik~P Kingma and Jimmy Ba.
\newblock Adam: A method for stochastic optimization.
\newblock {\em arXiv preprint arXiv:1412.6980}, 2014.

\bibitem{kocabas2020vibe}
Muhammed Kocabas, Nikos Athanasiou, and Michael~J Black.
\newblock Vibe: Video inference for human body pose and shape estimation.
\newblock In {\em CVPR}, 2020.

\bibitem{Kolotouros2019SPIN}
Nikos Kolotouros, Georgios Pavlakos, Michael~J Black, and Kostas Daniilidis.
\newblock Learning to reconstruct 3d human pose and shape via model-fitting in
  the loop.
\newblock In {\em Proceedings of the IEEE/CVF International Conference on
  Computer Vision}, pages 2252--2261, 2019.

\bibitem{kolotouros2019cmr}
Nikos Kolotouros, Georgios Pavlakos, and Kostas Daniilidis.
\newblock Convolutional mesh regression for single-image human shape
  reconstruction.
\newblock In {\em CVPR}, 2019.

\bibitem{prohmr}
Nikos Kolotouros, Georgios Pavlakos, Dinesh Jayaraman, and Kostas Daniilidis.
\newblock Probabilistic modeling for human mesh recovery.
\newblock In {\em Proceedings of the IEEE/CVF International Conference on
  Computer Vision}, pages 11605--11614, 2021.

\bibitem{hybrik}
Jiefeng Li, Chao Xu, Zhicun Chen, Siyuan Bian, Lixin Yang, and Cewu Lu.
\newblock Hybrik: A hybrid analytical-neural inverse kinematics solution for 3d
  human pose and shape estimation.
\newblock In {\em Proceedings of the IEEE/CVF Conference on Computer Vision and
  Pattern Recognition}, pages 3383--3393, 2021.

\bibitem{lin2021metro}
Kevin Lin, Lijuan Wang, and Zicheng Liu.
\newblock End-to-end human pose and mesh reconstruction with transformers.
\newblock In {\em Proceedings of the IEEE/CVF Conference on Computer Vision and
  Pattern Recognition}, pages 1954--1963, 2021.

\bibitem{lin2021_mesh_graphormer}
Kevin Lin, Lijuan Wang, and Zicheng Liu.
\newblock Mesh graphormer.
\newblock In {\em ICCV}, 2021.

\bibitem{lin2017feature}
Tsung-Yi Lin, Piotr Dollar, Ross Girshick, Kaiming He, Bharath Hariharan, and
  Serge Belongie.
\newblock Feature pyramid networks for object detection.
\newblock In {\em Proceedings of the IEEE conference on computer vision and
  pattern recognition}, pages 2117--2125, 2017.

\bibitem{lin2014mscoco}
Tsung-Yi Lin, Michael Maire, Serge Belongie, James Hays, Pietro Perona, Deva
  Ramanan, Piotr Doll{\'a}r, and C~Lawrence Zitnick.
\newblock Microsoft coco: Common objects in context.
\newblock In {\em European conference on computer vision}, pages 740--755.
  Springer, 2014.

\bibitem{liu2021Swin}
Ze Liu, Yutong Lin, Yue Cao, Han Hu, Yixuan Wei, Zheng Zhang, Stephen Lin, and
  Baining Guo.
\newblock Swin transformer: Hierarchical vision transformer using shifted
  windows.
\newblock {\em International Conference on Computer Vision (ICCV)}, 2021.

\bibitem{liu2021video}
Ze Liu, Jia Ning, Yue Cao, Yixuan Wei, Zheng Zhang, Stephen Lin, and Han Hu.
\newblock Video swin transformer.
\newblock {\em arXiv preprint arXiv:2106.13230}, 2021.

\bibitem{liu2021group}
Ze Liu, Zheng Zhang, Yue Cao, Han Hu, and Xin Tong.
\newblock Group-free 3d object detection via transformers.
\newblock In {\em Proceedings of the IEEE/CVF International Conference on
  Computer Vision}, pages 2949--2958, 2021.

\bibitem{SMPL:2015}
Matthew Loper, Naureen Mahmood, Javier Romero, Gerard Pons-Moll, and Michael~J.
  Black.
\newblock {SMPL}: A skinned multi-person linear model.
\newblock {\em ACM TOG}, 2015.

\bibitem{mpi3dhp2017}
Dushyant Mehta, Helge Rhodin, Dan Casas, Pascal Fua, Oleksandr Sotnychenko,
  Weipeng Xu, and Christian Theobalt.
\newblock Monocular 3d human pose estimation in the wild using improved cnn
  supervision.
\newblock In {\em 3D Vision (3DV), 2017 Fifth International Conference on}.
  IEEE, 2017.

\bibitem{misra2021end}
Ishan Misra, Rohit Girdhar, and Armand Joulin.
\newblock An end-to-end transformer model for 3d object detection.
\newblock In {\em Proceedings of the IEEE/CVF International Conference on
  Computer Vision}, pages 2906--2917, 2021.

\bibitem{Moon_I2L_MeshNet}
Gyeongsik Moon and Kyoung~Mu Lee.
\newblock I2l-meshnet: Image-to-lixel prediction network for accurate 3d human
  pose and mesh estimation from a single rgb image.
\newblock In {\em ECCV}, 2020.

\bibitem{PyTorch}
Adam Paszke, Sam Gross, Soumith Chintala, Gregory Chanan, Edward Yang, Zachary
  DeVito, Zeming Lin, Alban Desmaison, Luca Antiga, and Adam Lerer.
\newblock Automatic differentiation in pytorch.
\newblock 2017.

\bibitem{hmrsurvey}
Yating Tian, Hongwen Zhang, Yebin Liu, and Limin Wang.
\newblock Recovering 3d human mesh from monocular images: A survey.
\newblock {\em arXiv preprint arXiv:2203.01923}, 2022.

\bibitem{mlpmixer}
Ilya~O Tolstikhin, Neil Houlsby, Alexander Kolesnikov, Lucas Beyer, Xiaohua
  Zhai, Thomas Unterthiner, Jessica Yung, Andreas Steiner, Daniel Keysers,
  Jakob Uszkoreit, et~al.
\newblock Mlp-mixer: An all-mlp architecture for vision.
\newblock {\em Advances in Neural Information Processing Systems},
  34:24261--24272, 2021.

\bibitem{videomae}
Zhan Tong, Yibing Song, Jue Wang, and Limin Wang.
\newblock Video{MAE}: Masked autoencoders are data-efficient learners for
  self-supervised video pre-training.
\newblock In {\em Advances in Neural Information Processing Systems}, 2022.

\bibitem{resmlp}
Hugo Touvron, Piotr Bojanowski, Mathilde Caron, Matthieu Cord, Alaaeldin
  El-Nouby, Edouard Grave, Gautier Izacard, Armand Joulin, Gabriel Synnaeve,
  Jakob Verbeek, et~al.
\newblock Resmlp: Feedforward networks for image classification with
  data-efficient training.
\newblock {\em IEEE Transactions on Pattern Analysis and Machine Intelligence},
  2022.

\bibitem{deit}
Hugo Touvron, Matthieu Cord, Matthijs Douze, Francisco Massa, Alexandre
  Sablayrolles, and Herv{\'e} J{\'e}gou.
\newblock Training data-efficient image transformers \& distillation through
  attention.
\newblock In {\em International Conference on Machine Learning}, pages
  10347--10357. PMLR, 2021.

\bibitem{vaswani2017attention}
Ashish Vaswani, Noam Shazeer, Niki Parmar, Jakob Uszkoreit, Llion Jones,
  Aidan~N Gomez, {\L}ukasz Kaiser, and Illia Polosukhin.
\newblock Attention is all you need.
\newblock In {\em Advances in neural information processing systems}, pages
  5998--6008, 2017.

\bibitem{pw3d2018}
Timo von Marcard, Roberto Henschel, Michael Black, Bodo Rosenhahn, and Gerard
  Pons-Moll.
\newblock Recovering accurate 3d human pose in the wild using imus and a moving
  camera.
\newblock In {\em European Conference on Computer Vision (ECCV)}, sep 2018.

\bibitem{PVT}
Wenhai Wang, Enze Xie, Xiang Li, Deng-Ping Fan, Kaitao Song, Ding Liang, Tong
  Lu, Ping Luo, and Ling Shao.
\newblock Pyramid vision transformer: A versatile backbone for dense prediction
  without convolutions.
\newblock In {\em Proceedings of the IEEE/CVF International Conference on
  Computer Vision}, pages 568--578, 2021.

\bibitem{RSB-resnet}
Ross Wightman, Hugo Touvron, and Herv{\'e} J{\'e}gou.
\newblock Resnet strikes back: An improved training procedure in timm.
\newblock {\em arXiv preprint arXiv:2110.00476}, 2021.

\bibitem{xie2021segformer}
Enze Xie, Wenhai Wang, Zhiding Yu, Anima Anandkumar, Jose~M Alvarez, and Ping
  Luo.
\newblock Segformer: Simple and efficient design for semantic segmentation with
  transformers.
\newblock {\em Advances in Neural Information Processing Systems},
  34:12077--12090, 2021.

\bibitem{xu2020Low_Resolution}
Xiangyu Xu, Hao Chen, Francesc Moreno-Noguer, L{\'a}szl{\'o}~A Jeni, and
  Fernando De~la Torre.
\newblock 3d human shape and pose from a single low-resolution image with
  self-supervised learning.
\newblock In {\em ECCV}, 2020.

\bibitem{poolformer}
Weihao Yu, Mi Luo, Pan Zhou, Chenyang Si, Yichen Zhou, Xinchao Wang, Jiashi
  Feng, and Shuicheng Yan.
\newblock Metaformer is actually what you need for vision.
\newblock In {\em Proceedings of the IEEE/CVF Conference on Computer Vision and
  Pattern Recognition}, pages 10819--10829, 2022.

\bibitem{Thundr}
Mihai Zanfir, Andrei Zanfir, Eduard~Gabriel Bazavan, William~T Freeman, Rahul
  Sukthankar, and Cristian Sminchisescu.
\newblock Thundr: Transformer-based 3d human reconstruction with markers.
\newblock In {\em Proceedings of the IEEE/CVF International Conference on
  Computer Vision}, pages 12971--12980, 2021.

\bibitem{tcformer}
Wang Zeng, Sheng Jin, Wentao Liu, Chen Qian, Ping Luo, Wanli Ouyang, and
  Xiaogang Wang.
\newblock Not all tokens are equal: Human-centric visual analysis via token
  clustering transformer.
\newblock In {\em Proceedings of the IEEE/CVF Conference on Computer Vision and
  Pattern Recognition}, pages 11101--11111, 2022.

\bibitem{pymaf2021}
Hongwen Zhang, Yating Tian, Xinchi Zhou, Wanli Ouyang, Yebin Liu, Limin Wang,
  and Zhenan Sun.
\newblock Pymaf: 3d human pose and shape regression with pyramidal mesh
  alignment feedback loop.
\newblock In {\em Proceedings of the IEEE International Conference on Computer
  Vision}, 2021.

\bibitem{zhao2021graformer}
Weixi Zhao, Yunjie Tian, Qixiang Ye, Jianbin Jiao, and Weiqiang Wang.
\newblock Graformer: Graph convolution transformer for 3d pose estimation.
\newblock {\em arXiv preprint arXiv:2109.08364}, 2021.

\bibitem{gtrs}
Ce Zheng, Matias Mendieta, Pu Wang, Aidong Lu, and Chen Chen.
\newblock A lightweight graph transformer network for human mesh reconstruction
  from 2d human pose.
\newblock In {\em ACM Multimedia}, 2022.

\bibitem{Poseformer_2021_ICCV}
Ce Zheng, Sijie Zhu, Matias Mendieta, Taojiannan Yang, Chen Chen, and Zhengming
  Ding.
\newblock 3d human pose estimation with spatial and temporal transformers.
\newblock In {\em Proceedings of the IEEE/CVF International Conference on
  Computer Vision (ICCV)}, pages 11656--11665, October 2021.

\bibitem{zheng2021rethinking}
Sixiao Zheng, Jiachen Lu, Hengshuang Zhao, Xiatian Zhu, Zekun Luo, Yabiao Wang,
  Yanwei Fu, Jianfeng Feng, Tao Xiang, Philip~HS Torr, et~al.
\newblock Rethinking semantic segmentation from a sequence-to-sequence
  perspective with transformers.
\newblock In {\em Proceedings of the IEEE/CVF Conference on Computer Vision and
  Pattern Recognition}, pages 6881--6890, 2021.

\bibitem{zhu2020deformable}
Xizhou Zhu, Weijie Su, Lewei Lu, Bin Li, Xiaogang Wang, and Jifeng Dai.
\newblock Deformable detr: Deformable transformers for end-to-end object
  detection.
\newblock {\em arXiv preprint arXiv:2010.04159}, 2020.

\bibitem{freihand}
Christian Zimmermann, Duygu Ceylan, Jimei Yang, Bryan Russell, Max Argus, and
  Thomas Brox.
\newblock Freihand: A dataset for markerless capture of hand pose and shape
  from single rgb images.
\newblock In {\em Proceedings of the IEEE/CVF International Conference on
  Computer Vision}, pages 813--822, 2019.

\end{thebibliography}
}


\end{document}